\documentclass[runningheads,a4paper]{llncs}

\usepackage[numbers]{natbib}
\usepackage{amsmath}
\usepackage{amssymb}
\usepackage{longtable}
\usepackage{caption}
\usepackage{dcolumn}
\usepackage{graphics}
\usepackage{supertabular,booktabs}
\usepackage{graphicx}

\usepackage{enumerate}
\usepackage{caption}
\newcolumntype{d}[1]{D{.}{.}{#1}}

\usepackage{url}
\urldef{\mailsa}\path|{alfred.hofmann, ursula.barth, ingrid.haas, frank.holzwarth,|
	\urldef{\mailsb}\path|anna.kramer, leonie.kunz, christine.reiss, nicole.sator,|
	\urldef{\mailsc}\path|erika.siebert-cole, peter.strasser, lncs}@springer.com|    
\newcommand{\keywords}[1]{\par\addvspace\baselineskip
	\noindent\keywordname\enspace\ignorespaces#1}

\begin{document}
	%\begin{frontmatter}                           % The preamble begins here.
		\mainmatter  % start of an individual contribution
		
		% first the title is needed
		%\title{Rule Axioms in Ontologies: A review of the Semantic Web Rule Language (SWRL) Expressiveness Extensions}
		\title{The Semantic Web Rule Language Expressiveness Extensions --- A Survey}
		
		% a short form should be given in case it is too long for the running head
		\titlerunning{SWRL Expressiveness Extensions--- A Survey}

\author{Abba Lawan\inst{1}\textsuperscript{,} \inst{3} \and Abdur Rakib\inst{2}}% \and Natasha Alechina\inst{3} }

\authorrunning{A. Lawan and A. Rakib}

\institute{School of Computer Science\\ The University of Nottingham, Malaysia Campus\\ 
	\email{khyx3alw@nottingham.edu.my}
	\and
	Department of Computer Science and Creative Technologies\\
	The University of the West of England, Bristol, UK\\
	\email{Rakib.Abdur@uwe.ac.uk}
%	\and
%	School of Computer Science\\
%	The University of Nottingham, UK\\
%	\email{Natasha.Alechina@nottingham.ac.uk}
	\and
	Crops For the Future (CFF), Malaysia\\ 
}

\toctitle{Lecture Notes in Computer Science}
\tocauthor{Authors' Instructions}
\maketitle

		\begin{abstract}
			%% Text of abstract
		The Semantic Web Rule Language (SWRL) is a direct extension of OWL 2 DL with a subset of RuleML, and it is designed to be the rule language of the Semantic Web. This paper explores the state-of-the-art of SWRL's expressiveness extensions proposed over time. As a motivation, the effectiveness of the SWRL/OWL combination in modeling domain facts is discussed while some of the common expressive limitations of the combination are also highlighted. The paper then classifies and presents the relevant language extensions of the SWRL and their added expressive powers to the original SWRL definition. Furthermore, it provides a comparative analysis of the syntax and semantics of the proposed extensions. In conclusion, the decidability requirement and usability of each expressiveness extension are evaluated towards an efficient inclusion into the OWL ontologies.
	\end{abstract}
		% and the We further discuss the decidability requirements of the reviewed SWRL extensions in relation to the basic DL-safety restriction of SWRL rules. We conclude the paper with our observations on the usability of SWRL rules and a table of summary categorizing the various extensions, which are mainly: the fuzzy and probabilistic extensions for managing uncertainties, the non-monotonic extensions and existentials, advanced mathematic extensions, and some notable built-in extensions for added expressiveness to the classical SWRL definition.

		
	%\end{frontmatter}
	
	%% \linenumbers
	
	%% main text
	\section{Introduction}
	\label{intro}
	
	The OWL/SWRL combination offers a more flexible ontology language for modeling knowledge domains with higher degree of expressiveness than using OWL alone. However, despite the high expressive powers added by SWRL, the combination does not guarantee a comprehensive ontology modeling language. SWRL is found to be lacking descriptive constructs to model complex real-life scenarios as required by the Semantic Web. This has led to the evolution of various expressiveness extensions to the classical SWRL formalism, ranging from simple mathematical built-ins to extensions that facilitate modeling of vague facts and predictive knowledge of a given domain.
	
	\par
\begin{figure}[h]
		\centering
		\includegraphics[width=6cm,height=4.5cm]{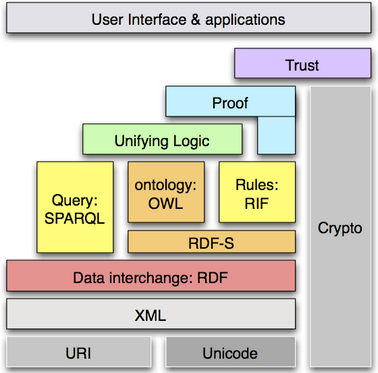}
		\caption{The Semantic web language stack\cite{Berners-Lee2009}}
		\label{fig:swebstack}
	\end{figure}

 The semantic web rule language (SWRL) \cite{horrocks2004swrl} is a W3C recommendation that extends the Web Ontology Language (OWL)\cite{mcguinness2004owl} with horn-clause rules. This allows declarative representation of complex domain information that may not be possible in OWL alone. OWL being the recommended ontology language for the semantic web \cite{TimBerners-Lee2011}, has shown considerable expressive powers over other ontology languages, especially its predecessor, the Resource Description Framework (RDF)\footnote{https://www.w3.org/RDF/}. However, while OWL ontologies provide simple, reusable and easy to understand domain knowledge models, they lack the declarative expressiveness offered by rules. As evidently shown in the semantic web architecture (Fig.\ref{fig:swebstack}), Rules are projected to support ontologies for efficient domain knowledge representation and the semantic web rule language (SWRL) is one of such rules' languages syntactically closest to OWL. Expressive limitations of the RDFS and OWL formalisms can thus be augmented by the capabilities of rules designed for the complex assertion of facts that goes beyond simple declaration of domain concepts.  Moreover, being in common practice with well-established logics, rules such as SWRL, offers an efficient reasoning support to ontologies with of course the benefit of added expressiveness.
	
	\subsection {The SWRL Formalism}
	SWRL is an expressive rule language designed to enable declarative assertions using OWL concepts. It is fashioned as the union of Horn logic (HL) and Description logic (DL) in order to achieve higher domain expressiveness and reasoning capacity than when using OWL alone. In other words, SWRL is a direct extension of OWL that utilizes its model-theoretic semantics and its syntax basically stemmed from the combination of DL-based OWL and HL-based Rule-ML. Classical SWRL rules include positive, function-free horn-clauses written as implications --- consisting of an antecedent (body of the rule), as well as consequent(or head of the rule). A SWRL-enabled ontology thus contains an OWL Knowledge base and set of horn-clause rule axioms.
	
	\subsubsection{SWRL Syntax and Semantics}
	In its human readable form, both SWRL's rule body (B) and rule head (H) typically consist of a conjunction of atoms, which can contain a combination of OWL constructs.  Such constructs can be either in the form of OWL-DL class descriptions of the form C(x), individual-valued properties --- \textit{P(x,y)}, data-valued properties --- \textit{Q(x,y}), OWL same individuals --- \textit{sameAs(x,y)}, OWL different individuals --- \textit{differentFrom(x,y)}, or the specific built-in functions --- \textit{builtIn(r, x, ...)}. Where 'x' and 'y' are either variables representing OWL classes, properties, and individual data values (in which case preceded by a '?') or OWL individuals themselves. \newline  Note: 'r' is any SWRL built-in function --- bound to a dataLiteral, such as: \textit{swrlb:greaterThan(), swrlb:Multiply()} etc.
	\newline \par
	\textit{ atom $B_{1} \wedge atom B_{2} ... \wedge atom B_{n} \longrightarrow  atom H_{1} \wedge atom H_{2} ...  \wedge atom H_{n}$ } \quad (1).
	\label{eq1}
	\newline \par The semantics ensures that the condition in the rule's head be evaluated to true whenever the conditions in the body are satisfied. However, there are variants to the basic format above such as rules with disjunctions in their heads or those extended to handle classical negations, etc. The aim of this paper is to review such various expressiveness extensions attributed to the SWRL formalism.
	
	\subsubsection{Decidability of SWRL Rules}
	\label{SWRLdecidability}
	
	\par Even though SWRL offers an unrestricted opportunity of combining ontologies and declarative assertion of rules, it however, does so at the expense of decidability. As clearly pointed out in \cite{Grigoris2008}, "there is undoubtedly no inference engine that can draw exactly the same conclusion as the SWRL semantics". This is due to the highly expressive nature of the formalism and SWRL being in first-order horn clause is unlike its complement OWL-DL, undecidable. Therefore, for SWRL ontologies to be decidable, a restriction needs to be placed on the rule axioms. To achieve decidability, SWRL rules are made to conform to \textit{DL-Safety}, which is a name for a restriction imposed on the rule axioms such that they contain only known concepts \cite{Rosati2005b}. In essence, variables in DL-safe SWRL atoms must be bound only to those concepts or individuals that are known to exist in the ontology. DL-based reasoners (such as Pellet, HermiT and Fact++, etc) are needed to reason over SWRL rules. \par
	While the DL-safety may result in an incomplete expression of facts in a given ontology (due to the limits imposed on the language), inferences from DL-safe rules are always formally sound. However, as is the case with other logic programs, such efforts to keep the SWRL formalism within decidable language constructs and computationally feasible inferences, have resulted in considerable limitations to the ontology rule language. This is evidently discussed in section \ref{SWRLlimits}, and thus leading to various expressiveness extensions of SWRL proposed over time. In what follows, we briefly compare the A-Box and T-Box decidability requirements followed by highlights in section \ref{whySWRL} on the strategic importance of SWRL in OWL ontologies --- a reasonable justification for our study. For brevity, reasoning paradigm in SWRL is not discussed here and we refer the interested reader to \cite{mei2005reasoning}.
	\newline \newline
	\emph{T-Box versus A-BOX Decidability:} Depending on the inference-subject involved, the DL-safety can be interpreted in terms of the T-Box (Concepts or Terminology Box) or the A-Box (Individuals or Facts Assertion Box) of a given ontology. In terms of the T-Box, DL-Safety imposes a limitation that \textit{only those classes or terms previously defined in the ontology can be used in the SWRL rules} --- meaning that no new or unknown individual may be added to the terminology box. Whereas in relation to the A-Box, the DL-Safety can be interpreted thus: \textit{variables used in the consequent of a rule must have also appeared in its antecedent} --- which implies that no anonymous or loosed individuals can be introduced into the ontology's A-Box. \par 
	Understanding these limitations can thus make it easier to design new SWRL extensions that stay within the decidable fragment of the DL, thereby allowing the inference engines or Reasoners to complete inferences within a finite time. Where such design cannot be easily achieved, a syntactic limitation or manual restriction were usually imposed on the use of the new extensions as discussed in details in section \ref{decidabilityreq}.
	
	\subsection{Why SWRL?}
	\label{whySWRL}
	Rules generally allow efficient declarative assertions in domain knowledge modeling. However, the semantic web rule language (SWRL) is particularly important in the semantic web knowledge modeling as it allows both the assertion of facts in OWL ontologies as well as their retrieval. Information retrieval is achieved through its slightly modified Semantic Web Query language(SQWRL), which is an SQL-extended version of SWRL for querying OWL ontologies \cite{O'Connor:2009:SQWRL}. The flexibility for users to define  application-specific methods (user-defined built-ins) as extension to the SWRL formalism is an important feature that makes SWRL formalism indispensable where domain modeling using OWL is considered. Regarding the added expressiveness of SWRL to the OWL language, the authors of \cite{Horrocks:2005:ORP} summarize some of the basic advantages of SWRL as follows: 
	\begin{enumerate}[i.]
		\item The ability to use class names or their descriptions as predicates. \item The use of equalities and inequalities, and \item Allowing conjunctions of atoms in both SWRL's antecedent and consequent. 
	\end{enumerate}
	
	These syntactic advances, coupled with the numerous expressiveness advantages of logic programming rules, enables SWRL to achieve complex representation of domain knowledge --- making the formalism indispensable in the Semantic Web project. 
	Further reasons for our focus on the SWRL formalism include the facts that:
	\begin{enumerate}[i.]
		
		\setcounter{enumi}{3}
		\item SWRL Rules can be used in transferring characteristics from one class or property to another without sub-classing --- which goes beyond the expressive powers of OWL. A commonly cited example of this transferring property is the composite property assertion, popularly referred to as 'the uncle relationship'.
		\item SWRL rules allow inference of new individuals using existential operators (e.g. \textit{swrlx: makeOWLThing}), some of which are defined as built-ins in the formalism (see section \ref{swrlbuiltins}). While the built-in semantics is yet to be formally defined as a SWRL specification, the ability of SWRL to achieve existential quantification as well as its syntax extensions through user-defined built-ins, goes beyond the expressive powers of OWL and the classical horn clause rule formalisms. 
		
		\item Lastly, SWRL being a semantic extension of OWL have enjoyed considerable experts' commitment and engaging support from the Semantic Web research community. \end{enumerate}
	Consequently, the compatibility of the SWRL formalism with OWL and also the RuleML, makes it easier and sometimes even compelling, for researchers interested in semantic web rules to equally extend SWRL's capabilities whenever the corresponding OWL language is augmented with additional syntax and semantics. A similar view was upheld in \cite{Calero2010}, where the authors affirms that the added expressiveness of OWL to cover negative property assertions inspired them to consider a corresponding SWRL extension. SWRL's negation extensions were first described and analyzed in \cite{patel2004proposal}. 
	
	\subsection{Limitations of SWRL}
	\label{SWRLlimits}
	Despite the above discussed expressive powers obtained from SWRL, the combination (of OWL and SWRL) still does not guarantee an all-inclusive domain modeling language. Classical SWRL formalism cannot appropriately represent various real-world scenarios. For example, expert opinions, which forms a considerable part of every domain knowledge, are typically in the form of imprecise facts. Thus requiring formalisms that are capable of representing domain facts based on some \textit{degree of certainty} or partial truth. Likewise, in legacy software models, much of formalized domain facts, such as business rules may only hold where other fragments of existing knowledge remain valid. Therefore, accurate representation of such scenarios requires \textit{knowledge exclusion, prioritization of facts,} as well as \textit{knowledge retraction,} among others. These forms of knowledge modeling scenarios are crucial to the effectiveness of expert systems and their related applications. With common examples found in information fusion, multimedia information processing, automated ontology merging and alignment, among others. In the remainder of this subsection, however, we highlight some of the commonly cited limitations of the classical SWRL formalism as follows:
\begin{enumerate}[i.]
		\item Limitation in modeling imprecise domain knowledge. The evolutionary and sometimes inconsistent nature of human knowledge necessitates representing vague or imprecise domain information. Hence, the inability of SWRL to represent inherently vague domain knowledge and business uncertainties is therefore, a huge setback in modeling real-world scenarios on the semantic web. This has led to the various fuzzy and probabilistic extensions of the semantic web rule language.
		\newline
		\item	Lack of Non-monotonic constructs. Rule-based modeling of a knowledge domain involves expressing domain facts or situations of domain objects, through the use of 'if-then' statements. This ordered representation of fact, basically reflects the ordered nature of the human knowledge --- which typically includes ordering or setting a precedence of activities, leading to the addition of new facts to existing knowledge bases. The inability of SWRL to comprehensively model non-monotonic facts such as existentials, quantifiers, rule exclusion, and prioritization, etc., were also considered as weighty limitations that cannot be otherwise ignored. Moreover, it has been shown that deductive forms of inference, where new facts get inferred from the absence of other facts, cannot be represented by the SWRL formalism \cite{Eiter:2008:CAS:1379469.1379712}.
			\newline
		\item	Removal of Facts. Further examples of realistic scenarios encountered include the need for 'unlearning' facts in knowledge bases, i.e. removal of facts. Constructs needed to model and achieve such scenarios were found to be lacking in the classical SWRL definition. 
			\newline
		\item	Lack of Support for Modeling Complex Scientific Knowledge. Other limitations include the SWRL's inability to efficiently model scientific and engineering knowledge domains, especially those involving complex mathematical formulas and constraints \cite{Kifer2005reqs}. The lack of precise constructs to handle complex engineering formulas usually leads to the development of voluminous set of rules to explain few facts.
	\end{enumerate}

	\subsection{Motivation and Scope}
	The limitations of OWL and SWRL formalism to provide some of the needed constructs to model complex scenarios are a hindrance in achieving accurate representation of domain knowledge on the semantic web. The aim of this work, therefore, is partly to explore the existing approaches proposed to eliminate these limitations. We evaluate the various expressiveness extensions to the SWRL formalism. To the best of our knowledge, there is no existing work that reviews the different SWRL expressiveness extensions. We consider that the work presented in this paper can help early researchers and developers in the field of semantic technologies to identify the versatility as well as scope of the SWRL/OWL combination. Moreover, the expressiveness extensions discussed can serve as a guide for future enhancement and standardization of SWRL, OWL, and their related formalisms.
	We limit our explorative review of the available literature to published (though not necessarily peer-reviewed) works on SWRL expressiveness extensions. While some of these proposals are largely theoretical, there are exceptional cases where the proposed extensions were supported with justifiable implementations. Our comparison of these extensions is largely analytical with little efforts to verify all the implementation details for the extensions. Moreover, as the review is intended to serve as a quick comparison guide to the SWRL expressiveness extensions, the discussions do not cover all aspects presented in the original research works. Our focus is mainly on their added syntax and semantics, implementation efficiency, followed by brief notes on their decidability requirements.
	\newline
	\par In the remainder of the paper, we set the stage in the next subsection with the Preliminaries, highlighting the basic terminologies and their interpretations within the context of this research. Essence of the paper is presented in Section \ref{SWRLexpressivenessext}, where we discuss and summarize the various extensions of the SWRL formalism. The decidability requirements of the reviewed extensions is expounded in section \ref{decidabilityreq}. This is followed by a discussion and summary in Section \ref{discussionandsummary} and lastly, conclude the paper in Section \ref{conclusion} highlighting some future research works. 
	
	\subsection{Basic Terms and Interpretations}
\begin{figure*}[t]
	\includegraphics[width=\textwidth,height=8cm]{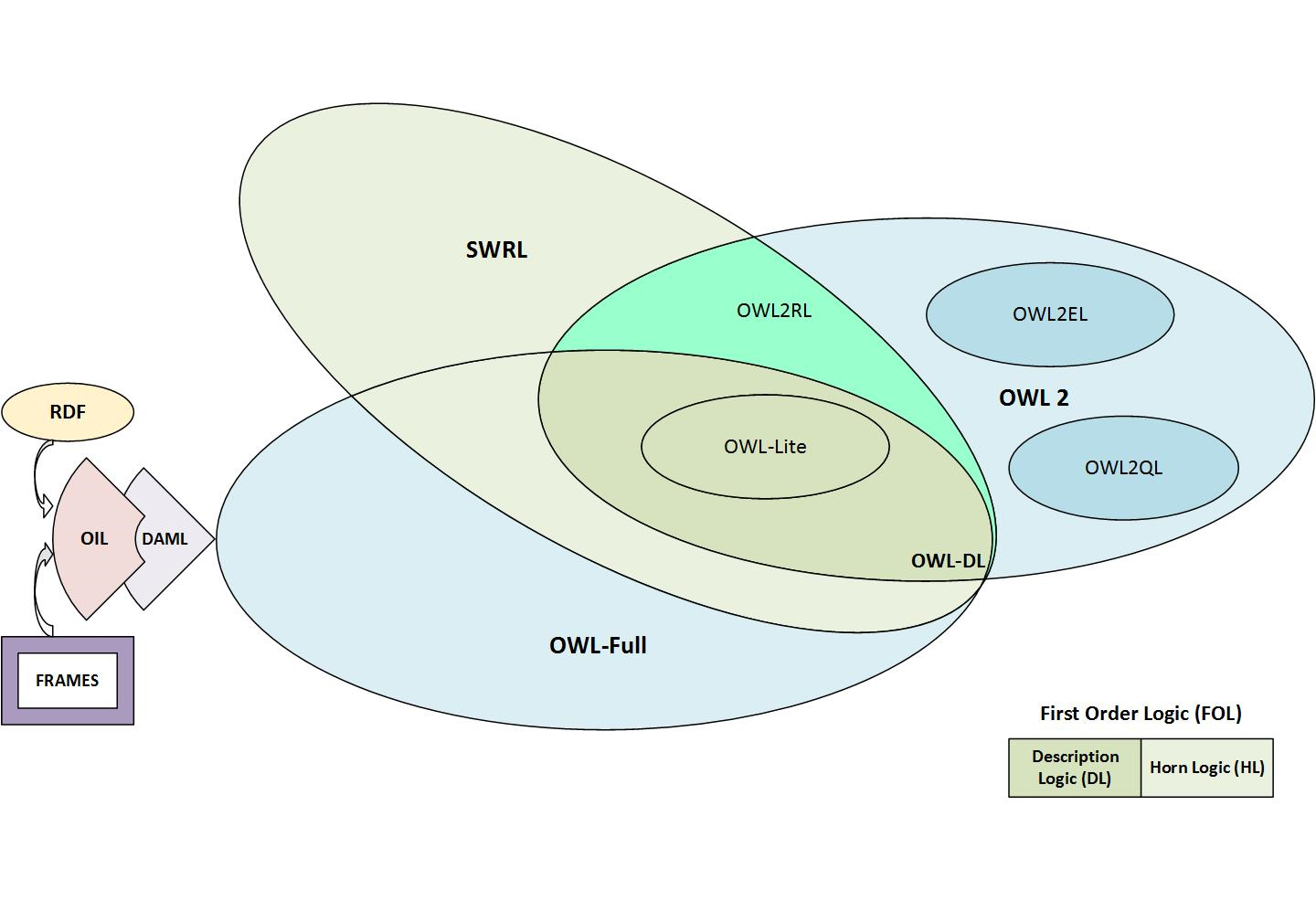}
	\caption[OWL Evolution and Contextual Relationship with SWRL]{OWL Evolution and Contextual Relationship with SWRL}
	\label{fig:owlevolution}
\end{figure*}

	\begin{itemize}
		\item Clausal Logic (CL) is a particular kind of First-Order Logic (FOL) --- a powerful formalism used to express relationships between objects in a given knowledge domain through quantified variables and predicates. CL provides the basis for Logic Programming (LP) through its fragment called Horn Clause Logic (HCL). 
		\item A Horn-Clause is logic program rule containing disjunctions of literals (rule atoms) with at most one positive literal \cite{horn1951onsentences}. A finite collection of Horn clauses together with ground facts are referred to as logic programs (LP). Simply put, LPs are a finite set of rules and facts. 
		\item Datalogs on the other hand, are function-free horn-clauses, having only variables and constants as terms. 
		\item Description Logic (DL) \cite{Baader2003DLH1} is basically a decidable fragment of FOL that allows modeling of knowledge domains using concepts (classes) the binary relations between them (called roles or properties), and individual instances (facts). Being decidable, the DL forms the basis of most domain modeling (ontology) languages including OWL.
		
		\item Decidability: In the context of this work, decidability refers to the ability of a 'Reasoner' to achieve inference --- checking the consistency of logical consequences of ontology axioms and return true or otherwise, within a finite time. That is, given any Ontology, a set of rules, and a sentence, the Reasoner can check that the sentence is entailed by the ontology and rules. More importantly, there exists a terminating procedure.
		
		\item OWL-DL is a decidable fragment of the Web Ontology Language (OWL). OWL being the most expressive ontology language for the Semantic Web, as recommended by the W3C \cite{mcguinness2004owl}, evolves overtime as improvement from other languages such as the RDF, FRAMES, DAML and OIL ontology languages (see Fig. \ref{fig:owlevolution}: OWL Evolution and Contextual Relationship with SWRL). Other fragments of OWL are the OWL-Full which subsumes the OWL-DL in terms of expressivity and the OWL-Lite, which is the least expressive fragment. To achieve reasoning over OWL ontologies, a subset of OWL-Full that completely conforms to the DL framework is extracted as the OWL-DL. Generally, OWL ontologies are restricted to the DL expressivity degree in order to achieve decidability and a complete OWL-DL ontology can have the expressive equivalence of $\mathcal{SHOIN(D)}$ in the DL expressiveness metric. 
		
		\item The DL Expressiveness Metric: In DL-based languages, factual knowledge during  terminology definitions and individual assertions are usually stored as formulas in First Order Logic (FOL). However, restrictions are usually attached to these formulas to ensure the decidability and for efficient reasoning over the ontology they represent \cite{shen2007using}. These restrictions also specify a degree of expressiveness of DL language as compared to the FOL, which is highly expressive --- though undecidable. Basic DL restrictions are represented by letter-symbol keys e.g. ($\mathcal{ALC, SHIQ, SHOIN(D),}$) etc, to denote a given expressive power of a modeling language, as described below:
		\begin{itemize}
			\item $\mathcal{S}$ \textemdash  An abbreviation of an Attributive Language with Complement $\mathcal{(ALC)}$ extended with transitive roles. 
			\item $\mathcal{ALC}$ --- this is obtained when AL is extended with the atomic concept negation such as for example, the Top concept $(\top \equiv C \sqcup \neg C)$ and Floor concept $(\bot \equiv C \cap \neg C)$, where C is an atomic concept. In owl, the Top concept is called 'Thing' \textit{owl:Thing} and all classes are subclasses of \textit{owl:Thing}, while the floor concept is considered as 'Nothing'.
			\item $\mathcal{AL}$ --- Attributive Language: is the basic DL language that allows the use of Concept intersection ($\cap$), universal restrictions ($\forall$), limited existential quantification ($\exists$) and atomic negations of concepts ($\neg$), which do not appear on the left-hand-side of axioms.
			\item $\mathcal{H}$ ---An abbreviation of $\mathcal{ALC}$ extended with the role hierarchy (owl: subPropertyOf relationship).
			\item $\mathcal{O}$ --- An abbreviation of $\mathcal{ALC}$ extended with Nominals (enumerated classes e.g. owl:oneOf or object value restrictions such as owl:hasValue relationship).
			\item $\mathcal{I}$--- An abbreviation of $\mathcal{ALC}$ extended with Inverse roles or properties, which allow expressing relationships in opposite directions (e.g. owl:hasPart and owl:isPartOf).
			\item $\mathcal{N}$--- An abbreviation of $\mathcal{ALC}$ extended with a Number or cardinality restriction. Semantics:  $\geq$ n R.C or $\leq$ n R.C
			\item $\mathcal{D}$ ---The data values expressivity ($\mathcal{D}$), which is sometimes attached to the algorithm as subscript, denotes the abilities of DL and its family of languages to use data values, datatype, and datatype properties to further express domain facts.\newline
		\end{itemize} 
	
	\item The OWL 2 Profile:
	\label{owl2}
	The Web Ontology Language version 2 (OWL 2) is the formally recommended ontology language for Semantic Web modeling and much like its predecessor OWL 1, it allows logical domain modeling by defining classes, individuals, their properties and relationships with each other or data values, with the addition that OWL 2 ontologies are exclusively stored as Semantic Web documents. Specifically, the current version OWL 2 is able to provide a wider range of constructs for expressing concepts such as $transitive$ and $inverse$ properties, $cardinality\ restrictions$, as well as $inheritance$, among others.  By targeting specific modeling needs of the web, it is thence divided into three sub profiles, viz. the 'OWL 2 Expressive Language' (OWL2EL), 'OWL 2 Query Language' (OWL2QL), and the 'OWL 2 Rules Language' (OWL2RL) ~\cite{Motik2009, Boris2012}. These sub-languages offer different expressiveness and computational desirability.	
	
\end{itemize}
	In essence, the web ontology language (OWL-DL) gives a very expressive and decidable ontology language and since OWL-DL utilizes the extensive research on DL, it serves as the foundation for the current language profiles of OWL 2. Fig. \ref{fig:owlevolution} below summarizes the contextual relationship between the OWL profiles and SWRL.
	
	%\lipsum[1-2]
%\lipsum[3-10]

	\section{The SWRL Expressiveness Extensions}
	\label{SWRLexpressivenessext}
	
	SWRL extensions are originally defined in \cite{HORROCKS2005} as \textit{bindings} that provide a mapping between variables used in the SWRL rules to objects of a given domain. As previously mentioned much of SWRL's extensions were inspired by their corresponding OWL extensions. This is usually achieved by modifying the OWL's abstract syntax and semantics to adapt to the SWRL canonic mode. This leads to various proposals that are beneficial to the advancement of both the OWL and SWRL languages.  The remainder of this section highlights some of the expressiveness extensions designed to address the commonly encountered limitations of the classical SWRL.
	
	\subsection{Fuzzy and Probabilistic Extensions --- Dealing with Uncertainties and Imprecise Knowledge}
	Knowledge from domain experts is usually not without doubts and imperfections. This imperfect nature of real-world information processing necessitates the long-started efforts by the AI community to deal with vague knowledge representation. Notable early efforts in the field of classical domain modeling have attempted to improve the expressivity of Description Logic (DL) to cover uncertainties through Fuzzy Logic (FL) as discussed in \cite{Wlodarczyk2010} and probabilistic extensions \cite{Predoiu2007}, among others. Likewise, SWRL is found to be inadequate in expressing practical problems that involve vague or imprecise domain knowledge – which is bound to be common in ontologies or the Semantic Web as a whole. Moreover, even with the success of semantic web in data provisioning, writing business and engineering rules using SWRL still remains a challenge. This is due to the fact that SWRL, much like other Semantic Web languages, stemmed from classical logics (see Figure \ref{fig:owlevolution}), which are known to be incapable of modeling vague or imprecise information. 
	\par
	As highlighted in \cite{Pan2006fswrl}, other scenarios that require modeling of imprecise facts can be found in multimedia processing, ontology alignment, and information fusion, among numerous others. In response, SWRL has since received significant advancements to handle imprecise domain knowledge. One famous uncertainty reasoning techniques involve the combination of fuzzy logic with SWRL and another involves extending SWRL formalism with probability theories. This section reviews some of these advancements and briefly discusses their individual approaches.
	
	\subsubsection{The Fuzzy-SWRL Extension (F-SWRL)}
	\label{fswrl}
	\paragraph{Expressiveness:} By imposing fuzzysets theory --- precisely, the 'R-implication' of fuzzy logics \cite{Klir1994fuzzysets}, new semantics were defined by Horrocks et al. for a fuzzy extension of SWRL (f-SWRL) \cite{Pan2006fswrl}. A fuzzy set is defined by its degree of individual membership (w) called weighted-degree or truth-value and usually computed from a \textit{membership distribution function}. The fuzzy membership function does not only specify whether an element belongs to a given set or otherwise but also how much. Analogous to fuzzy sets, f-SWRL rules uses a truth value between '0 and 1' to express the degree of confidence for individual membership in a given class or property and also to express weights or importance of each atom in a SWRL rule. 
	\paragraph{Syntax and Semantics:} Class and property definitions in f-SWRL have the following form $C(x) * w$ and $P(x,y) * w$, respectively. While a rule in f-SWRL is of the form:

	\textit{Antecedent} * $w_a$ $\longrightarrow$ \textit{Consequent} * $w_c. \quad$ Where $w \in [0, 1]$.
	\newline
	For example, the following f-SWRL rule (retained from \cite{Pan2006fswrl}), declares that "being healthy is more important than being rich to determine if one is happy":\par 
	\begin{equation}
	\label{rule1}
	Rich (?p) * 0.6 \wedge Healthy (?p) * 0.8 \longrightarrow Happy (?p) * 0.9
	\end{equation}

	Where Rich, Healthy, and Happy are fuzzy class URI refs, ?p is an individual valued variable and 0.6, 0.8 and 0.9 are the assigned weights of the corresponding fuzzy atoms. \newline \newline
	In a nutshell, f-SWRL extends SWRL with fuzzy-based class and property definitions as well as fuzzy rule axioms. The result is that f-SWRL rules are able to represent such information that says how much a concept is believed to be true and which facts are more important than others when making decisions. Users can, therefore, represent vague domain knowledge with f-SWRL rules by asserting the degree of confidence or otherwise of a given fact.
	
	\paragraph{Implementation and Efficiency:} In assigning a predetermined weight on the consequent atom, the \textit{f-SWRL} extension is shown to be suitable for writing fuzzy rules with atomic consequents and thereby making rule prioritization. This absolute form of prioritization, however, becomes a problem whenever new rules are introduced in the KB, as then all the existing weight values may have to be readjusted. Moreover, the proposal does not present a way of resolving such conflicts. As such, more support is needed for f-SWRL to deal with non-atomic fuzzy rules, i.e. rules with more than one atom in the consequent. Furthermore, complete fuzzy-rule declaration needs to be considered in f-SWRL since representing uncertainties in domain knowledge requires more than just fuzzy class and property definitions. 
	Based on our evaluation, much of the f-SWRL proposal is still theoretical as neither implementation efforts nor practical scenarios were mentioned. Likewise, the fuzziness of the extension has been critically questioned by the authors of \cite{agarwal2005modeling}, asserting that syntax and semantics proposed in f-SWRL do not actually solve much of the fuzzification problem. For an abstract literature on fuzzysets theory and fuzzy logics, we refer the interested reader to \cite{Klir1994fuzzysets}.
	
	\subsubsection{The Vague-SWRL Extension}
	\label{vagueSWRL}
	
	Highlighting their argument against f-SWRL's inability to provide a substantial fuzzy extension to SWRL, the Vague-SWRL extension was proposed in \cite{Wang2008}. 
	
	\paragraph{Expressiveness:} In this proposal, the authors argues that the use of single-membership degree on a fuzzy set, such as the single weight function in f-SWRL extension, is insufficient in representing a vague information. Consequently, based on the theory of Vague sets \cite{gau1993vague}, the authors propose the Vague-SWRL as another fuzzy extension of SWRL. Vague Sets are themselves an extension of fuzzy sets, where the degree of membership to a set is evaluated using two weighted intervals --- as opposed to a single degree of membership employed in fuzzy sets. As such, Vague-SWRL rules uses an added weight value ($w_2$) called 'a second-degree weight' denoting further degree of membership to support and balance an initial weight ($w_1$). By introducing $w_2$, the proposal promises a more accurate representation of imprecise domain knowledge than a single membership fuzzy class and properties assertions of f-SWRL. 
	
	\paragraph{Syntax and Semantics:} As presented in \cite{gau1993vague}, the general form of vague-SWRL is written as:
	
	\begin{flushright}
		\textit{(vc * fdw) (vcv * sdw) ...} $\wedge$ \textit{(vp * fdw) (vpv * sdw) ...} $\longrightarrow$ \textit{(vc * w) or (vp * w)}  
	\end{flushright}
	
	Where: 'vc' = vague classes, 'fdw' = first-degree weights, 'vcv' = vague class values corresponding to 'vc' and 'sdw' = second-degree weights. Similarly, 'vp' = vague properties, 'vpv' = vague property values and 'w' = the atomic weights. 
	The vague values, also called the membership intervals, are calculated as:
	\begin{equation*}
	vcv/vpv = [t_v (x), \quad 1 - f_v(x)]
	\end{equation*} 
	
	And the 'sdw', also referred to as the 'vagueness' or second degree of membership, is calculated as the difference:
	\begin{equation*}
	w_2 = [(1-f_v(x)) - t_v (x)]
	\end{equation*} 
	Where: $t_v$ (x) is the true membership function of x, $f_v$(x) its false membership function, and $ 0 \leq t_v (x) + f_v(x) \leq 1$
	
	\paragraph{Example Case.} Consider the following additional information (membership degrees) added to rule (1): \textit{"P is rich with a true value of 0.6 and a false value of 0.3. Also, P is healthy with a true value of 0.3 and a false value of 0.2"}. 
	\par With this additional info, we can represent the parameters:
	\begin{equation*}
	vcv(Rich) = [0.6, 0.7]\ and\ vpv(isHealthy) = [0.3, 0.8]
	\end{equation*}
	Hence, the vague-SWRL form of rule (1) can then be written thus:
	\begin{equation}
	\begin{split}
\label{rule2}
[Rich (?p) * 0.6] [0.3] \wedge [Healthy (?P) * 0.8] [0.1]\\ \longrightarrow Happy (?p) * 0.9
	\end{split}
	\end{equation}

	Here, 0.9 is the degree to which the consequent holds following the evaluation of the antecedent. 
	
	\paragraph{Implementation and Efficiency:} As vague sets subsumes fuzzy sets, vague-SWRL ultimately subsume its corresponding f-SWRL rule in terms of expressiveness. By comparing rules \ref{rule1} and \ref{rule2}, it can be seen that unlike f-SWRL, vague-SWRL is more than just a conjunction of weighted atoms. The added effort in the form of $vcv$ and $vpv$ are calculated to represent the uncertainties in the fuzzy membership classes Rich, Healthy, and Happy. Moreover, by specifying the upper and lower bounds of membership intervals (through the true and false values), vague-SWRL rules are more justifiably accurate in representing imprecise knowledge using fuzzy class and properties. Vague-SWRL is also claimed, by the authors, to be in "an acceptable form of the Rule Interchange Format (RIF)". A comparable extension, \textit{Vague-RuleML}\cite{Paschke2010}, is also proposed for the rule markup language (Rule ML). 
	\par However, similar to f-SWRL, the Vague-SWRL only represents fuzzy information for class and property memberships, which may be inadequate in representing practical uncertainties involved in knowledge domains and the semantic web. Also, a fair understanding of vague sets and vague knowledge representation is required to efficiently model fuzzy information using vague-SWRL rules. Hence, there is a need to improve the proposal with richer fuzzy modeling syntaxes to handle imprecise domain knowledge beyond the borders of class and property memberships --- the SWRL-F extension below gives a good example.
	
	\subsubsection{The SWRL-Fuzzy Extension (SWRL-F)}
	\label{swrlf}
	The SWRL-F extension \cite{Wlodarczyk2010} was proposed as another Fuzzy Logic (FL) extension to the SWRL formalism. However, unlike f-SWRL and vague-SWRL extensions, which were based on the fuzzy sets principle, the SWRL-F extension expresses fuzzy reasoning in SWRL rules using a 'fuzzy control system' approach.
	\paragraph{Expressiveness:}  In this approach, the main ontology remains intact while a fuzzy ontology consisting of SWRL-F rule-base is added to model any ambiguous domain knowledge using logical variables. The fuzzy ontology is needed to define inherent fuzzy domain knowledge using fuzzy: sets, terms, variables and fuzzy values as entities. These entities were defined as classes with their respective object and data properties. Modelling a fuzzy fact requires the use of the fuzzy terms to calculate a fuzzymatch for the respective instances.
	
	\paragraph{Syntax and Semantics:} SWRL-F introduces a special object property called 'fuzzymatch' for each fuzzy set. The fuzzymatch is used when designing SWRL-F rules to match corresponding Fuzzy-Variables with designated Fuzzy-Values from each respective fuzzy set. For example, the following SWRL-F rule (adopted from \cite{Wlodarczyk2010}) can assert the vague facts; \textit{"Persons with good health status are always very happy"} and can be written using SWRL-F rule as: 
	\begin{equation}
	\begin{split}
	Person (?p)\ \wedge\ hasHealthStatus (?p, ?s)\ \wedge\ \\fuzzymatch (?s, goodHealthStatus)
	\wedge\ isHappy (?p, ?h) \\ \longrightarrow fuzzymatch (?h, VeryHappy)
	\end{split}
	\end{equation}
	
	Here, the fuzzymatch attribute is used in the antecedent of the rule to calculate the degree of membership for the 'HealthStatus' variable (?s) --- as employed in the fuzzy term 'goodHealthStatus' denoting the fuzzy set in this instance. While used in the Consequent, the fuzzymatch variable allows binding the new value 'VeryHappy' to the fuzzy value 'isHappy', giving SWRL a chance to express the wooly term, 'very happy'. 
	
	\paragraph{Implementation and Efficiency:} In essence, the fuzzy ontology and SWRL-F rules are only added where necessary to handle uncertain domain knowledge representation. Furthermore, the authors claim that the new rule language, SWRL-F, is supported with an exclusive publicly-available  ontology development environment having a test execution engine. However, several attempts at finding the link currently present an empty Prot\'{e}g\'{e} wiki page. An apparent limitation to the SWRL-F implementation is that some of the current OWL2 constructs cannot be utilized in the SWRL-F rules. Moreover, running the SWRL-F rules requires a modified version of the \textit{SWRLJessTab} in Prot\'{e}g\'{e} ontology editor --- an implementation requirement that may be difficult to domain experts. Moreover, using the language to model vague domain knowledge also requires an in-depth understanding of fuzzy logic and representation scheme.
	\par On the other hand, due to its adoption of a well-established approach --- the fuzzy control system approach, SWRL-F offers a more pragmatic and justifiable approach to fuzzy extension of SWRL as compared to its siblings, f-SWRL and vague-SWRL. The use of the 'fuzzymatch' attribute also makes SWRL-F suitable for developing semantic web applications with queries that require fuzzy inferencing from facts separately stored in the main ontology. In addition, since by design the SWRL-F fuzzy ontology is separated from the domain ontology, reasoner inferences are thus limited only to the SWRL-F rulebase. This is desirable nonetheless, as the modularity avoids introducing inconsistencies to the main ontology.

	\subsubsection{Fuzzy Non-monotonic Extension of SWRL (f-NSWRL)}
	\label{fuzzynonmonotonic}
	
	The f-NSWRL \cite{Wangfnswrl}, presents yet another fuzzy SWRL extension for dealing with non-monotonicity as well as uncertainties in domain information using SWRL rules.
	\paragraph{Expressiveness:}  This extension is basically an advancement of the fuzzy-SWRL extension (section \ref{fswrl}) to incorporate a non-monotonic knowledge, specifically the knowledge negation (called classical negation) and Negation as Failure (NAF), which involves knowledge modeling based on the absence of positive facts in a KB. This is also referred to as Closed World Negation (CWN). Citing the importance of expressing classical negation and the NAF in handling rule exceptions, the authors asserts the motivations for f-NSWRL extension of SWRL --- as a non-monotonic as well as a fuzzy extension of the SWRL formalism. 
	
	\paragraph{Syntax and Semantics:} In its simplest form, the f-NSWRL uses the 'Not' and '$\neg$' symbols as operators to extend the original f-SWRL's fuzzy classes and properties definition. As an example, consider the addition of the following information to modify rule \ref{rule1} in an attempt to determine if a person is happy; \textit{"Person p is rich and p is definitely not hungry but the health status of p is not known"} Decision rules based on the above statement can be expressed using f-NSWRL as follows:
	\begin{equation}
	\label{rule4}
	\begin{split}
	Healthy (?p) * 0.8 \wedge (\neg Hungry (?p)) * 0.99\\ \longrightarrow Happy (?p) * 0.9
	\end{split}
	\end{equation}
	
	\begin{equation}
	\label{rule5}
	\begin{split}
	Healthy (?p) * 0.8 \wedge not(Hungry (?p)) * 0.99\\ \longrightarrow Happy (?p) * 0.9
	\end{split}
	\end{equation}	
	
	Note the difference in the use of the two operators as highlighted in the two equations. While the classical negation ($\neg$) is used to negate the assertion that Person (?p) is hungry in rule \ref{rule4} and the NAF used in \ref{rule5} to test the absence of an assertion that 'p is hungry'. A quick interpretation of the two rules is that rule \ref{rule4} simply declares that 'A person must be \textit{not hungry and healthy} to be happy'. Whereas rule \ref{rule5} declares that 'a person is happy if she is healthy and is \textit{not known} to be hungry'. 
	
	\paragraph{Implementation and Efficiency:} Apart from handling negation and uncertainty, other notable aspects of the f-NSWRL proposal include a proposed markups in 'RuleML' for translating f-NSWRL to other rule languages and also procedures to handle rule prioritization --- for setting rule precedence in cases of conflicting consequents. However, the proposal failed to mention the semantics of these extensions nor the inference mechanism for implementing a supporting Reasoner.
	\par We discuss more on specific Non-monotonic extensions of SWRL in section \ref{nonmonotonic} below.
	
	\subsubsection{The Bayesian Extension of SWRL (Bayes-SWRL)}
	\label{bayesian}
	
	In order to allow modeling of predictive knowledge and the representation of inherently probabilistic domain knowledge (such as Statistical information) on the semantic web, various probabilistic extensions were proposed --- to the semantic web languages. These include among others, the Probabilistic RDF (pRDF) and Probabilistic OWL (PR-OWL) \cite{DaCostaPROWL}, Bayes-OWL\cite{Ding2006} and Bayes-SWRL \cite{Liu2013}. 
	
	\paragraph{Expressiveness:}  While Fuzzy Logic extensions of SWRL focus on representing the degree of certainty or otherwise of domain knowledge in SWRL-enabled ontologies, probabilistic extensions of SWRL such as Bayes-SWRL are more concerned with representing predictive knowledge based on an existing partial knowledge in the domain ontology or knowledgebase. Based on the Bayesian Networks of probability theory \cite{Darwiche2010} and in line with the corresponding Bayesian OWL extension \cite{Ding2006}, the Bayes-SWRL extends the SWRL formalism with the ability to model probabilistic knowledge. This is achieved by combining SWRL with the expressive capabilities of the Bayesian Logic Programs thereby enabling probability assertions during inference.
	
	\paragraph{Syntax and Semantics:} The Bayes-SWRL extension naturally adopts the original SWRL's abstract syntax and semantics with the addition of few innovative symbols and patterns to represent the probability variable (the p-variable) and related terms --- such as the Conditional Probability Table (CPT). Similar to the fuzzy weights (w) assigned to Vague-SWRL rule atoms, the Bayes-SWRL also use a Probability variable to assign probabilities to rules atoms in both the antecedent and/or consequent atoms. The probability value also ranges from [0, 1] and is optionally added to the rule atoms, with its values predefined in a conditional probability table supplied to the consequent atom in the form of an XML file. In its human readable syntax, Probability values of imprecise atoms are attached with an asterisk (*) to the rule atoms. While the CTP file path is attached using the '@' symbol (as shown in rule \ref{cptrule} below). In this form, a Bayes-SWRL that asserts that \textit{"It might rains with a certain probability, if it is cloudy and humid"} can be written as:
	\begin{multline}
	\label{cptrule}
	Cloudy (?cl)\ *\ p(cl)\ \wedge\ isHumid (?hm)\ *\ p(hm)\longrightarrow\\ Rainfall(?r)\ *\ p(r)@"RainfallCPTs.xml"
	\end{multline}
	
	Where: the p-variables, * p(cl), * p(hm) and *p(r), represents the probability values for cloudy, humid and rainfall, respectively. Assuming probable rainfall chances were calculated, for optimal cloudy and humid conditions, in the supplied \textit{RainfallCPTs.xml} file. A more detailed description of the abstract syntax and semantics of Bayes-SWRL can be found in  \cite{Liu2013}.
	
	\paragraph{Implementation and Efficiency:} The proposal of Bayes-SWRL is equipped with a reasoning algorithm implemented by extending an existing OWL-DL reasoner based on the tableaux algorithm, the Pellet reasoner, to interpret the added syntax based on the defined probabilistic semantics. It is also equipped with a user interface for checking rules conflict and viewing the inference process, among others. \newline 
	
	Consider this: The extension is effective as it is able to provide a well-defined syntax and semantics for modeling uncertainties. However, the extension being a by-product of BLP and SWRL automatically inherits the constraints of both formalisms. Hence for Bayes-SWRL rules to be decidable, the DL-safety restriction must be adhered to by its reasoner. Similarly, the BLP preconditions apply to a Bayes-SWRL rule, such as: firstly, a consequent atom can only be influenced by a finite number of random variables --- thereby adopting the closed world assumption. While this may not pose any serious threat to domain modeling, it however, limits the capabilities of Bayes-SWRL to represent NAF. Secondly, in the inference relation between atoms, there can be no cycle in the dependency graph. Hence expressing relationships is severely limited. Thirdly, a probability of any consequent atom can only be influenced by the probabilities of corresponding antecedent atoms in the same rule. Meaning that, probabilities defined in previous rules cannot be reused in new rules. Therefore, contrary to the non-monotonic nature of the real-world knowledge, especially those found in the semantic web, this precondition demonstrates a monotonic limitation of Bayes-SWRL knowledge bases. Moreover, while the conditional probability of a Bayes-SWRL rule is based on a well-founded semantics, as is the case with pDatalogs \cite{Predoiu2007}, the declaration of an arbitrary set of probable states for the ground atoms beforehand, poses some close-world expressive limitation for modeling continuously evolving knowledge domains.
	
	\subsection{Non-monotonic Extensions --- Dealing with Negation and Removal of Facts, Rules Exclusion and Prioritization}
	\label{nonmonotonic}
	
	With the proposed extension of the OWL2 profile to handle negative property assertions, there is a corresponding effort to also extend SWRL with non-monotonic operations --- notably presented in \cite{patel2004proposal} and \cite{Calero2010}. Moreover, since SWRL formalism basically involves OWL constructs coupled with rule axioms, any OWL extension inherently results in a new SWRL extension. While non-monotonic axioms involve constructs that produce knowledge bases conforming to the Closed World Assumption (CWA), monotonic constructs generally follow the Open World Assumption (OWA), as required in the Semantic Web environment. As such, the semantics of these non-monotonic extensions to SWRL have to be carefully modeled to keep the new rules decidable. For brevity of scope, we present here only the non-monotonic extensions added to the SWRL formalism. For a detailed description of Monotonicity and the applications of non-monotonic logic in rules, we refer interested readers to \cite{Calero2010}. 
	
	\subsubsection{The 'not' operator --- Weak Negation or Negation as Failure (NAF)}
	
	Owing to its monotonic background inherited from Description Logics, and the OWA of ontologies, the original SWRL formalism does not support negation as failure (NAF)--- as that will logically violate the open-world protocol. As such, both its antecedent and consequent can contain only positive conjunctions of atoms or facts.
	\paragraph{Expressiveness:}  Hence, the NOT operator (also called weak negation) was introduced to allow for expressing negation of facts. Negation of facts simply means, modeling the absence of positive, known or existing facts in a knowledgebase. It should be noted here that whenever a negation of membership is intended, a strong negation (also called complement) is easily achieved in SWRL using the \textit{owl:complementOf} class description.
	
	\paragraph{Syntax and Semantics:} Abstract definition of the \textit{not} operator is depicted in the following table:
	
	\begin{table}[ht]
		\centering
		\begin{tabular}{p{2.0cm}|p{3.8cm}|p{2.1cm}}
			%{|c|c|c|}
			\hline 
			SWRL \newline Elements & Pattern Matching (P) & Condition Test \\ 
			\hline \hline 
			P(x; y)	\newline Q(x; y) & (?x P ?y)\newline (?x Q ?y)\newline & S(P) = True \\
			not(P(x; y))\newline not(Q(x; y)) & (?z rdf:type owl:Negative-Property) \newline
			(?z rdf:subject ?x)\newline
			(?z rdf:predicate [P or Q])\newline
			(?z rdf:object ?y) & S(P) = False
			
		\end{tabular} 
		\caption{Syntax of the Weak Negation (NAF) \cite{Calero2010}}
		\label{tab:weaknegation}
	\end{table}
	
	\emph{Example Case.} The following Non-monotonic SWRL rule declares that \textit{"A person that does not have a 'spouse property' is automatically a member of the Singles class"}:
	\begin{equation}
	\label{rule7}
	Person(?p) \ \wedge\ Not(hasSpouse(?p, ?s)) \longrightarrow Single(?p) 
	\end{equation}

	For the complement operator or Strong Negation (as previously discussed in f-NSWRL --- see section \ref{fuzzynonmonotonic}), using the above example, we can write the rule by testing the existence of 'P' in the married class, as follows:
	\begin{equation}
	\label{rule8}
	\begin{split}
	Person(?p) \ \wedge\ (notMarried(?p)) \longrightarrow Single(?p)
	\end{split} 
	\end{equation}
	Where, the 'not' enclosed in the bracket together with the 'Married' class name, denotes the \textit{complementOf} relationship for the married class.
	
	\paragraph{Implementation and Efficiency:} However, a special Reasoner is required to run extensions having the weak negation (NAF) as shown above. On how to express the semantics of NAF, the 'non-monotonic inference process' has been highlighted in \cite{Calero2010}. On the other hand, the classical or strong negation can be easily achieved in OWL and subsequently in SWRL, using the \textit{owl:complementOf} class description, which can also be employed in the SWRL rules  --- as shown in rule \ref{rule8}.
	
	\subsubsection{The Quantifiers --- Exists, ForAll, and notExists operators}
	
	The Quantifiers were introduced in SWRL rules to handle incomplete facts, when used in the antecedent (body of the rule) and for removing facts, when used in the rule's consequent (head). In this set of expressiveness extensions, two types of fact quantification were introduced here. First, the Existential quantifiers --- the Exists ($\exists$) and notExists($\nexists$) used to assert an existence and otherwise respectively, of at objects in a knowledge base (KB). Secondly, the Universal quantifier --- ForAll ($\forall$), which generalizes some assertion on group of objects in a KB. 
	\par We discuss the \textit{notExist} extension here. While the \textit{Exists} and \textit{ForAll} quantifiers with their example use cases, will be discussed in section \ref{existentials} as part of the SWRL existential extensions.
	
	\paragraph{Expressiveness of 'notExist' Quantifier} Due to the different semantic interpretations attributed to the antecedent and consequent sides of rule axioms, the notExist quantifiers can be used to achieve different assertion of facts (expressiveness) depending on which side they are placed in a SWRL rule. These assertions are presented in the following scenarios:
	
	\begin{enumerate}[a.]
		\item \textit{notExists} Operator in the Antecedent --- Checking for missing facts:
		To handle missing information as facts, the 'notExists' and 'Exist' quantifiers were proposed to allow some action to be taken where certain facts are not defined in the KB. This is useful because due to the OWA of SWRL KBs, it is not always feasible to write rules that, for example, enumerate or test-out for all individuals or properties. In such cases, asking for the existence or otherwise of a particular individual or property offers a simple solution. For example, the following rule (\ref{rule9}) checks for the absence of participants with spouses in a booking register and asserts the status of the register.
		\begin{multline}
		\label{rule9}
		Participant(?p)\ \wedge\ hasBooking (?p, ?b)\ \wedge\\ \textbf{notExists}(hasSpouse(?p, ?s)) \longrightarrow\\ bookingStatus (?b, "Singles Only")
		\end{multline}
		\item \textit{notExists} Operator in the Consequent --- Removal of Knowledge:
		Conversely, using the NotExist operator in the consequent of a SWRL rule, results in the removal of knowledge. For example, rule \ref{rule9} above may need to be retracted whenever a married participant made a booking and the 'hasSpouse' property get added to the ontology. Instead of deleting the rule manually, a better option, would be to change the 'bookingStatus' information using another rule (\ref{rule10}), as follows:
		\begin{multline}
		\label{rule10}
		Participant(?p)\ \wedge\ hasBooking (?p, ?b)\ \wedge\\ hasSpouse(?p, ?s) \longrightarrow\\ \textbf{notExists} (bookingStatus (?b, "Singles Only"))\\ \wedge\ bookingStatus (?b, "Mixed")
		\end{multline}
		
	\end{enumerate}
	
	A clearer example is also presented in the mail list update function shown in rule \ref{rule11}, where the 'notExist' operator is used to test whether a member of a workgroup exist in a membership mailing list before adding them to an Alumni mail list:
	\begin{multline}
	\label{rule11}
	Workgroup(?g)\ \wedge\ Alumni(?a)\ \wedge\ hasMember(?g,\\ ?a) \longrightarrow \textbf{notExists}(mailListMember (?a, ?g)\\ \wedge\ AlumniMailList(?a)
	\end{multline}
	
	\paragraph{Syntax and Semantics of the 'notExist' Quantifier}
	In continuation of the definition of the Non-monotonic operators, the Table \ref{tab:notExists} below shows the notExists constructs and is used as is, in the human-readable syntax of SWRL. The semantics hold that S(P) becomes universally satisfiable for any unbound elements (A, B ... Z) associated to the notExist operator.
	\begin{table}[ht]
		\centering
		\begin{tabular}{p{3.0cm}|p{2.6cm}|p{1.5cm}}
			%{|c|c|c|}
			\hline 
			SWRL Elements & Pattern Matching (P) & Condition Test \\ 
			\hline \hline 
			NotExist (A, B, ... Z) & 	 (A), (B), ..., (Z)  & S(P) = U  
			
		\end{tabular} 
		\caption{Syntax of 'notExist' Extension to SWRL \cite{Calero2010}}
		\label{tab:notExists}
	\end{table}
	
	\paragraph{Implementation and Efficiency:} As shown in the example rules \ref{rule9}, \ref{rule10} and \ref{rule11}, implementation of the \textit{notExists} operator is simple and usually depends on the side of the inference operator it appears. A notable efficient use of the operator is where there are conflicting facts in a knowledge base, the use of \textit{notExists} operator help in retraction of a rule to restore consistency. As such, the concept of rule's 'precedence' and 'rule retraction' is strongly advised in the inference process of \cite{Calero2010}. Consequently, this paves the way for our next SWRL extension --- the \textit{dominance} and \textit{mutex} operators in the next section.
	
	\subsection{Rules Ordering and Priority Extensions}
	Introduced to define relationships between rules and their semantics, the 'dominance' and 'mutex' constructs are extensions meant to control the behavior of SWRL rules execution.
	
	\paragraph{Expressiveness:}   Precisely, the dominance operator is used to specify the order of rule execution by assigning a precedence of one rule over another. In contrast, the 'mutex' operator is non-symmetric and designed to assert the complete exclusion of a rule due to the execution of other rules.
	
	\paragraph{Syntax and Semantics:} As defined in \cite{Calero2010}, the dominance operator is defined as \textit{dominance ($R_x$, $R_y$)}, where $R_x$ and $R_y$ are rule names, and with a semantic meaning that "rule $R_x$ has more priority in the execution order than $R_y$". Furthermore, the dominance operator is designed to be transitive, thereby suitable for handling the addition of new rules into a rule base.  \par
	As an example, consider the execution of rules \ref{rule9} and \ref{rule10} above. We may want to run the 'Mixed-Status' rule (rule \ref{rule10}) first to check if there are married participants in the knowledge base before asserting the 'Singles' booking status in rule \ref{rule9}. To set this ordering, we can write the dominance operation as follows:
	\begin{equation}
	dominance\ (Rule_{10},\ Rule_9)
	\end{equation}
	
	Whereas in its syntactic format, the mutex operator is written thus: \textit{Mutex ($R_x,\ R_y$)}, asserting the fact that "rule $R_y$ will not be executed in the event that rule $R_x$ is already executed". Continuing to the hypothetical example above, we may want to completely skip executing the 'Singles Only' rule (\ref{rule9}) in the event that the booking status is explicitly known to be 'Mixed'. This can be achieved that through thus:
	\begin{equation}
	Mutex\ (Rule_{10},\ Rule_9)
	\end{equation}
	
	\paragraph{Implementation and Efficiency:} However, these extensions requires specially modified Reasoners for inference and are yet to be formally accepted as part of standardized SWRL definition.
	
	\subsection{Existential Extensions --- Dealing with Quantification of Individuals}
	\label{existentials}
	As explained earlier, owing to DL-Safety restrictions, SWRL-enabled ontologies can not introduce new individuals (see section \ref{SWRLdecidability}). Moreover, with SWRL being the combination of OWL-DL and DataLog RuleML --- which do not allow existential quantification in its consequent, it becomes even more difficult to assert new individuals into a classical OWL/SWRL ontology. However, as the authors of \cite{Li2011} puts out, \begin{quote}
		"It sometimes becomes necessary during inference and where certain conditions are met, to introduce new individuals to an ontology"
	\end{quote}
	Addition of new individuals to a knowledge base is commonly referred to as 'existential quantification' and in what follows, we discuss the various SWRL extensions proposed for achieving that.

	\subsubsection{The XSWRL Extension}
	\label{xswrl}
	
	\paragraph{Expressiveness:} To achieve existential quantification in SWRL, new operators were defined and the new extension referred to as extended semantic web rule language (XSWRL) \cite{Li2011}. The direct extension defines new operators to achieve the addition of new individuals to existing classes of SWRL ontologies. 
	
	\paragraph{Syntax and Semantics:} XSWRL uses similar syntax and semantics of the classical SWRL with the only difference being that rules in XSWRL use an additional operator that allows for introducing new individuals. This is achieved through the use of 'existentially quantified variables' in the rule's consequent --- represented by prefixing them with an exclamation mark (!). Universally quantified variables are represented with the usual question mark (?), as in the original SWRL definition. As an example, consider the rule assertion that \textit{"All members of a project Workgroup should be Engineers"}.
	\begin{equation}
	\label{rule14}
	Workgroup (?g)\longrightarrow Engineer(!e)\ \wedge\ hasMember(?g, ?e)
	\end{equation}
	
	\paragraph{Implementation and Efficiency:} From the XSWRL's prototype implementation, presented in \cite{Li2011}, it can be understood that by defining the semantics of '!' (the existential operator) into the SWRL abstract semantics, reasoning over XSWRL rules can be achieved using existing DL-Reasoners such as Racer and Fact++. However, it remains to be seen of such implementation and whether the extension will be practically utilized by the semantic web community. This is because: contrary to the classical SWRL, XSWRL rules with non-atomic consequents and having joint existential variables cannot be split into multiple rules with atomic consequents. This, makes it difficult to assign rule-preference or prioritize rule execution of XSWRL rules.
	Recall, that the concept of DL-Safeness ensures decidability in SWRL rules by limiting the consequent variables to only those that previously occurs in the antecedent. As such, further restrictions must be placed on the use of the existential variables declared in XSWRL to ensure decidability. To this end, the authors declare the added restriction: \textit{"Do not construct a rule, which has existentially quantified variables, to form acyclic chain between its atoms or with atoms from other rules"}. Though this may restrict the expressiveness of the XSWRL language, but it ensures that infinite chains are safely avoided. Hence, XSWRL rules need to be tracked manually to ensure that their executions will not lead to a cyclic chain of existential quantification. However, this manual restriction, can be overly tedious or impracticable in very large ontologies.
	
	\subsubsection{SWRL First Order Logic Extension (SWRL-FOL)}
	\label{swrlfol}
	\paragraph{Expressiveness:} In an attempt to extend SWRL towards the expressiveness of First-Order Logic, notably to achieve the quantification of individuals, a SWRL-FOL extension was proposed in \cite{patel2005proposal}. Analogous to the FOL Rule Markup Language \footnote{http://ruleml.org/fol/}, the proposal defines an abstract syntax and semantics for the SWRL-FOL extension. It shows how SWRL can be extended, to utilize the expressive powers of FOL, by extending the component OWL axioms to include function-free FOL assertion axioms. 
	\paragraph{Syntax and Semantics:} SWRL FOL proposal defines an abstract syntax of the expressive extensions and further provided the model theoretical semantics for their interpretations. In their abstract syntax, SWRL-FOL ontologies contain sets of OWL axioms, facts, and horn-clause rules with additional FOL axioms or assertions. These assertions however, introduces some extensions to the original SWRL format, such as the limitless use of 'conjunctions' and/or 'disjunctions' in the FOL formula and the use of constructs such as \textit{negation, 'ForAll', 'Exists'}, etc. over unary and binary predicates. However, the semantic interpretations of these assertions are defined as 'bindings' which maps every variable to an element in the domain. See SWRL built-ins extensions in section \ref{swrlbuiltins} for more details on predicate bindings.
	
	\emph{Example:} Consider the Workgroup members assertion in rule \ref{rule14}. An alternative expression using the direct \textit{Exists} element defined in FOL-SWRL can be written as: 
	\begin{equation}
	\label{rule15}
	Workgroup (?g) \longrightarrow \exists_m hasMember (?g, ?m)\ \wedge\ Engineer (?m)
	\end{equation}
	
	\paragraph{Implementation and Efficiency:} What we noted here is that the proposal in \cite{patel2005proposal} is largely theoretical and as the authors of \cite{Li2011} puts out, its practical implementation is still open for discussion. This however, is due to the usual concerns of decidability and the lack of Reasoners that can achieve inferences over the FOL sentences. Nevertheless, considering its extensive definition and the general utilization of FOLs, no doubt the SWRL-FOL extension can help to provide a good foundational framework for FOL-based SWRL extensions.
	
	\subsubsection{SWRL Constraints Interchange Format (CIF-SWRL): Extending SWRL to Express Fully Quantified Constraints}
	\label{cifswrl}
	
	\paragraph{Expressiveness:} In their motivation, the authors of CIF-SWRL \cite{McKenzie2004a} explains that knowledge fusion in an open distributed environments such as the semantic web, involves data gathering from various network sources, which also include the constraints on how the data can be used. As such, utilizing these constraints directly using the SWRL rules can help to achieve existential quantifications. \newline
	The Constraint Interchange Format (CIF) extension of SWRL (CIF-SWRL) is an advancement of the SWRL formalism towards the constraint satisfaction problem (CSP). While in most cases, the target is simply to improve the SWRL formalism (rule layer) for better domain modeling, in CIF-SWRL extension, the objective is to improve SWRL to handle CSPs in the Logic layer of the Semantic Web and the aim is to allow the quantification of these constraints so that new individuals can be introduced into the knowledge base.
	
	\paragraph{Syntax and Semantics:} To allow expressing fully-quantified constraints using the CIF-SWRL rules, the proposal aligns the original features of CIF with the SWRL definition to form a single modeling language. In essence, the extension introduce constraints --- defined as quantified implications, to solve the problem of existential quantification in the classical SWRL. These constraints are expressed using an interchange format, such as the First-order logic (FOL)-based CIF, to form relevant constraint satisfaction problems that serve as inputs to constraint solver. To illustrate the CIF/SWRL added syntax, we quote the following rule assertion: \textit{"Every workgroup must contain at least 1 member who is a Professor"}
	\begin{equation}
	\begin{split}
	(\forall ?g \in Workgroup) \longrightarrow \\(\exists ?p \in Professor) \wedge hasMember(?g, ?p)
	\end{split}\end{equation}
	
	Meaning: For all Workgroup g, there exists a Professor p, who is also a member of the workgroup. The rule above shows the directly added symbols, 'ForAll ($\forall$) and Exists ($\exists$), which are both subsets of a 'Quantifiers' class --- defined in the CIF-SWRL expressiveness extension. 
	
	\paragraph{Implementation and Efficiency:} The CIF-SWRL extension comes about as an advancement of an existing CSP, which uses OWL ontology as a data model and SWRL rules for expressing constraints.  Beside the use of quantification symbols, CIF/SWRL also claims to introduce nested quantified implications, which supposedly allows for multiple-quantification of individuals in a rule. The proposal features the technical details of CIF-SWRL extension, including the abstract syntax and semantics, with an illustrative application to a use case. However, the featured implementation does not fully explain a reasoning strategy for inference on CIF-SWRL rules. Other forms of non-monotonic extensions, precisely the representation of 'disjunction' and 'negation', were also mentioned as work-in-progress.
		\par
	Note: Rule (16) can also be expressed simply in DL (OWL) as:
	\begin{equation}
	Workgroup\sqsupseteq \exists hasMember.Professor
	\end{equation}
	
	\paragraph{The Semantic Web Constraint Language (SWCL)} 
	Recognizing the need to improve the CIF/SWRL extension to handle more than just logical constraints, a complimentary extension of CIF-SWRL called the Semantic Web Constraint Language (SWCL) was proposed in \cite{Kim2003} and subsequently applied in \cite{kim2009semantic}. The rationale being that by incorporating other constraint satisfaction problems. such as the mathematical constraints, CIF-SWRL will can extend its applicability to problem areas beyond simple decision problems. The SWCL is then defined as an OWL-based extension for modeling mathematical constraints to solve basic optimization problems in the semantic web. It includes the proposal of a Unified logic and Constraint problem solver --- alleged to be a semantic-web-based decision-making framework for implementing case scenarios using the SWCL.
	
	\subsubsection{The Epsilon Existential Extension of SWRL}
	\paragraph{Expressiveness:} Proposed in \cite{Elenius}, the \textit{Epsilon} extension is another existential extension designed to allow quantification of individuals into OWL ontologies using SWRL rules. 
	
	 In this SWRL existential extension, the authors uses a new operator the Hilbert's Epsilon ($\epsilon$), to denote existential quantification. However, the main objective of the proposal is to achieve structural computation in OWL and subsequently, in the SWRL formalism. 
	\paragraph{Syntax and Semantics:} The epsilon operator is employed to define a 'terminology constructor' ($\epsilon_X \phi(X)$), which on inference, is expected to return anonymous individuals as 'values of class X' as results of an existential $Formula$ function $\exists_X:\phi(X)$. The syntax of the epsilon extension is defined by simply extending the abstract syntaxes of the 'i-objects' and 'd-objects' of SWRL to represent the epsilon terminology. While highlighting the limitations of procedural attachments such as SWRL built-ins, and the inefficiencies involved in using first order logic extensions, the semantics of the epsilon extension were described in \cite{Elenius} with theories and technical case study showing the use of the new operator. However, the discussion failed to show the usability of the new operator in SWRL rules and no supporting reasoners were mentioned. 
	
	\paragraph{Implementation and Efficiency:} In order to utilize the epsilon extension, an in-depth understanding of the theory of Hilbert's Epsilon operator seems to be inevitable. As no example SWRL rules were presented on how to utilize the new operator, implementation of the epsilon extension is still open for discussion. Moreover, the new operator seems to only increase the complexity of the SWRL formalism in achieving existential quantification when compared to the previous existential extensions discussed. 
	
	\par In addition, it should be noted that a turn-around fashion of creating new individuals in SWRL rules is possible with the use of the followings: (i) the OWL class construct \textit{owl:someValuesFrom} --- thereby achieving existential quantification as class assertions. (ii) Alternatively, the SWRL built-in \textit{swrlx: makeOWLThings} (discussed as part of SWRL Builtin extensions in section \ref{swrlbuiltins}), also allow direct creation of individuals in SWRL-enabled ontologies. Though, inference on the \textit{swrlx:makeOWLThings} built-in will result in free individuals that cannot be classified into the ontology. (iii) SWRL can also achieve existential quantification using the restriction \textit{owl:someValuesFrom} to directly creates new individuals into OWL classes. Even though the resulting quantification does conform to DL-safeness, such existential formulation using a restriction can be hard to implement in practice and has been criticized as being inconvenient. 
	
	\subsection{SWRL Extension for Advanced Mathematical Support}
	\label{mathextenstion}
	As a semantic web rule language, modeling knowledge for, and across, all types of domains should be possible using the SWRL formalism. This includes complex mathematical equations, typically used in engineering applications. While SWRL's mathematical built-ins support the basic arithmetic operations such as addition, subtraction, comparison, string and Boolean operations, etc. There is, therefore, a need for extending SWRL to handle complex mathematical and engineering computations such as polynomials, integration, differentiation, summation, etc. 
	
	\subsubsection{The OpenMath Extension of SWRL}
	\label{openmath}
	\paragraph{Expressiveness:} To this end, Lopez and others in \cite{sanchez2007extending} propose the combination of SWRL built-ins with the 'OpenMath'\footnote{http://www.openmath.org/} model to provide advanced mathematical support in SWRL. The OpenMath is an extensible representation standard as well as an evolving interchange framework for sharing and publishing mathematical objects and their semantics. It is basically, a markup language and a representation standard for mathematical objects. 
	
    The aim  of SWRL-Openmath extension is to allow expressing scientific knowledge using formulas. Using the functionalities of OpenMath, an additional SWRL built-in \textit{swrlbext:mathext} (with three basic arguments) was designed to extend the SWRL built-ins ontology class. In essence, the 'SWRL-OpenMath' extension extends the SWRL formalism with a feature that evaluates and reason over mathematical expressions. By using the built-in (swrlbext: mathext) to define functions as instances of a $Formula$ class, the OpenMath extension enables the representation of complex math operations such as integration, differentiation, polynomials, etc.
	
	\paragraph{Syntax and Semantics:} The mathematical expressions are represented using the OpenMath XML functions. The function parameters, which are assumed to be already defined in the OWL ontology, are then supplied as part of the SWRL rules. To allow reuse of the math functions, a $Formula$ class is defined to represent the OpenMath expressions as datatype values using a special datatype property called \textit{hasOMExpression}. The result of the expression i.e. the parameters supplied and the $Formula$ itself, comprises the three arguments defined in the built-in extension \textit{swrlbext:mathext} as described in the proposal.
	
	\paragraph{Implementation and Efficiency:} Apart from utilizing complex operators and their semantics, the use of OpenMath instead of the classical mathematical operators also helps to separate the mathematical and problem semantics in writing SWRL rules, thereby giving more clarity to rules representation. \par 
	However, the short proposal only gives an overview of the extension with a draft of a methodology showing how the combination can be achieved and a mention of possible implementation using 'Bossam' and 'Mathematica'. Moreover, the implementation details of the extension is unavailable for further analysis. It should be noted that, the OpenMath extension introduces a rather complex approach to handle formulas in SWRL as compared to the mathematical built-ins approach. In addition, the proposal failed to discuss reasoning supports for the OpenMath SWRL extension and is curiously silent on the overall decidability of the combination. 
	
	\subsection{SWRL Built-in Extensions --- Addressing SWRL limitations through Built-ins}
	\label{swrlbuiltins}
	
	Discussion on SWRL extensions can never be complete without mentioning the SWRL built-ins\footnote{http://www.daml.org/2004/04/swrl/builtins.html}. Simply put, SWRL built-ins are procedural attachments used to augment the expressive powers of the original SWRL language definitions. More formally, a SWRL built-in is defined as "a predicate that takes one or more variables as arguments and evaluates to true if the argument satisfies the predicate". SWRL built-ins consist of the 'core built-in libraries' for common operations involving constraints, lists, string, comparison, Boolean, URIs, and date operations --- preceded by the namespace qualifier \textit{swrlb:}. Moreover, SWRL built-ins are especially useful as they allow special definition of domain-specific, arbitrary methods called 'user-defined built-ins' --- an important feature of the SWRL formalism that allows users to define new built-in libraries for special tasks. Core built-ins such as the mathematical operators, and built-ins for string and date operations were defined in the original SWRL specification.
	
	\paragraph{Implementation:} Defining SWRL built-in extension is possible either directly in OWL or through their corresponding java implementations made possible by the \textit{SWRLBuiltInBbridge}. The \textit{SWRLBuiltInBbridge} is a component of the open-sourced SWRLTab of the Prot\'{e}g\'{e} ontology editor, which allows the manipulation of SWRL built-ins using Java. User-defined OWL-based built-in definitions can be achieved by simply adding them as new instances to the \textit{swrl:Builtin} class predefined in the SWRL definition ontology. Relevant built-ins are usually grouped together in a single OWL file, which can be imported into any domain ontology for utilization. A Java implementation of the built-ins, wrapped in a JAR file is however needed in the Prot\'{e}g\'{e}-OWL plugins directory for the Built-in bridge to make the necessary run-time linkages. A good example is the SWRL-Inference and Query tool, popularly termed as the SWRL-IQ --- originally defined as a "plugin for Prot\'{e}g\'{e} version 3.x that allows users to edit, save, and submit queries to an underlying inference engine based on XSB Prolog." The built-in extension (swrl-extension.owl) basically contains user-defined predicates, implemented for use in the SWRL Inference and Query Tool (SWRLIQ). \par 
	In what follows, we briefly review some of the most popular as well as standardized SWRL built-in extensions, such as the temporal built-ins, the mathematical built-ins, the semantic web query language (SQWRL), and the Existential built-ins of SWRL:
	
	\subsubsection{SWRL Temporal Built-ins}
	
	\paragraph{Expressiveness:} Due to the limited temporal support in both OWL and SWRL, another notable example in the SWRL expressiveness extensions is the SWRL Temporal Built-in Library \cite{Connor2011}.
	
    Defined as part of the SWRL-API's built-in library, the temporal built-ins are hierarchically defined in the SWRL temporal ontology. The SWRL temporal model is designed to allow easy representation of temporal knowledge in SWRL-based ontologies. The temporal ontology provides a standard model for modeling the temporal domain facts. As a result, the built-ins allow temporal reasoning on OWL ontologies using SWRL rules. 
	
	\paragraph{Syntax and Semantics:} The temporal built-ins provide a rich set of temporal operators such as \textit{before, after, during, duration, contains, overlaps,} etc. and are normally preceded by the name-space qualifier 'temporal:'. Based on the time data they operate, the SWRL temporal built-ins were categorized into basic and advanced mode.  In the basic mode, SWRL temporal built-ins operates on arguments supplied by the XML Schema's 'date' and 'dateTime' data types --- supplied as \textit{xsd:String} with values such as second, hour, day, time, week, month, year, etc. Whereas in the advanced mode, the SWRL temporal built-ins works on time information that is completely encoded using the 'valid-time' temporal model . \par 
	As an example, a rule that asserts the 'Fellow' membership rank by categorizing all registered Workgroup members with registration dates before the year 2000 can be written as: 
	\begin{multline}
	\label{rule17}
	WorkgroupMember(?m)\ \wedge\ hasRegDate(?m, ?rd)\\ \wedge\ \textbf{temporal}:before (?rd, '2000') \longrightarrow\\ FellowMembers(?m)	
	\end{multline}
	
	\subsubsection{SWRL-M Built-ins --- A Complex Mathematical Built-in Library.}
	\paragraph{Expressiveness:} Apart from the basic arithmetic operations available in the original SWRL definition, a SWRL-M built-in extension, was recommended to allow SWRL rules to handle more mathematical expressions. Defined as part of the SWRL Inference and Query Languages (SWRLIQ) \cite{elenius2012swrl}, the extension introduces complex math operations as an advancement to the core mathematical built-in library. It is efined as part of the SWRL Mathematical Ontology and written with the prefix \textit{swrlm:} as its pseudonym.
	\paragraph{Syntax and Semantics:} Common mathematical operations allowed in SWRLM built-in library include among others; the square root operation, \textit{swrlm:sqrt(?x, a)} and the evaluate expressive function, \textit{swrlm:eval(?x, "expression")}. The latter is designed to support the evaluation of the ontology variable x against standard constants and functions such as pi ($\pi$), epsilon ($\epsilon$), Lin ($\ln$), etc. For example, a SQWRL query that "returns a random number between 0 and 1" can be written as:
	\begin{equation}
	swrlm:eval (?x, "rand()") \longrightarrow sqwrl:select (?x)
	\end{equation}
	
	\paragraph{Efficiency:} The SWRL-M built-in provides a simple and efficient way of dealing with complex math operations --- especially when compared to the OpenMath extension (section \ref{openmath}). However, the simplicity also has its price as the SWRL-M built-ins collection falls short in representing many advanced mathematical expressions, some of which are highlighted in section \ref{mathextenstion}. \par 
	The restriction is a classical issue of modeling languages, i.e. the need to balance between tractability and degree of expressiveness. And as mathematical extensions involving recursive functions in rules usually makes inference non-terminating --- and hence entailment undecidable, there is the need for careful design of built-in functions to ensure that predicates introduced remain decidable.
	
	\subsubsection{SWRLX Built-ins --- The SWRL Existentials Built-in Library}
	
	\paragraph{Expressiveness:} Considering the need to create new individuals using the classical SWRL definition and without relying on external extensions, the SWRL existential built-in (\textit{swrlx:makeOWLThings}) is defined in \cite{rulemlswrlx}. It is defined in the SWRLX Ontology and written using the 'swrlx:' prefix. The built-in is designed to ease explorative modeling in SWRL. Specifically, the existential quantifications --- see section \ref{existentials}. Creating new individuals using SWRL built-in can be of particular importance, especially in the execution of mapping rules.
	\paragraph{Syntax and Semantics:} Using the SWRLX built-in model, creating new individuals is made possible using the \textit{swrlx:makeOWLThings} method and which has at least one free variable as its argument. The semantics being that the \textit{swrlx:makeOWLThings} will create a new individual of type \textit{owl:Thing} and binds it to the free variable in the method's argument. In essence, the axiom \textit{swrlx:make OWLThings}(?x, ?y) --- "will cause a new individual to be created and bounded to ?x for every value of the matching variable ?y in the rule". 
	
	Now consider an example SWRLX rule in (\ref{rule19}), which creates a new individual for every membership ID of the \textit{WorkgroupMembers} class and then asserts the new individual into the 'Editors' class.
	\begin{multline}
	\label{rule19}
	WorkgroupMember(?m)\ \wedge\ hasMemberID(?m, \\?mID) \wedge \textbf{swrlx:makeOWLThings}(?ed, ?mID) \\ \longrightarrow Editors(?ed)
	\end{multline}		
	
	\paragraph{Efficiency and Decidability:} Creating new individuals using SWRL built-in can be of particular importance, especially in the execution of mapping rules. The advantage of \textit{swrlx:makeOWLThings} method over other existential quantifications is that it offers a simple and direct method of creating new individuals in SWRL-based ontologies. However, as the new individuals created will simply be of type 'owl:Thing', it cannot be further classified by a reasoner into any particular class type. This may result in redundant unbounded individuals in the main ontology. As such an efficient and careful use of the built-in requires that a class be predefined in the ontology to collate the new individuals created. Another safe implementation, as the authors advised, will be to completely avoid storing the new individuals into the main ontology. However, SWRL built-ins still remains limited in extending the semantic web rule language to represent non-conventional domain knowledge. While those already defined, could benefit from well-documented constructs with more efficient syntax and semantics. \newline
	
	In conclusion, SWRL built-ins certainly increase the expressiveness of SWRL and the possibility of user-defined SWRL built-ins means that domain-specific extensions can always be defined to extend SWRL's expressive powers. Other built-ins not expanded in our discussion here, includes the query extension built-ins such as the OWL-Axioms: the T-Box, A-Box, and R-Box built-in libraries, which allows querying knowledge stored in the: terminology box, the assertions box and the Relations sets of OWL ontologies. Note: special query built-ins were grouped together to form the Semantic Query Web Rule Language (SQWRL) built-ins \cite{O'Connor:2009:SQWRL}. Thus to improve their usability, syntax of these built-in extensions need to be improved with well-defined semantics within the DL-safety restriction.
	
	\section{Decidability of SWRL Expressiveness Extensions}
	\label{decidabilityreq}
	
	From the foregoing review of the various expressiveness extensions added to SWRL, one may be interested in asking the question, "What then, could be the largest decidable extension of the semantic web rule language?" While an obvious answer could be the DL-Safe extensions, there is still the need to evaluate which assemblage of these extensions can still remain decidable. Where the combination is no longer DL-safe, then what restrictions could be imposed for the combination to remain decidable. This is particularly important in achieving inference and to ensure usability of these extensions. \par In practical terms, \textit{Decidability} refers to the ability of a Reasoner to classify and achieve inference over a given piece of ontology within a finite time. While basic SWRL rules can be kept decidable through the DL-Safety restriction, most of the SWRL expressiveness extensions can hardly be kept decidable by employing similar restriction. Such extensions that work within the fringes of decidability, such as the existential extensions, need to be carefully tailored to ensure that their usage do not cause inconsistencies to the resulting ontology. For example, the XSWRL extension \cite{Li2011}, which adds new individuals using existentially quantified variables is clearly non Dl-safe and therefore, undecidable. This is because, unbounded variables are bound to be introduced into the ontology, possibly creating anonymous individuals or cyclic chains in rule execution. As such for the XSWRL extension to be decidable, there must not be any chain of existentials (whether cyclic or acyclic) in rule executions.
	In what follows, we briefly discuss the conditions needed to maintain the decidability of all the reviewed SWRL extensions:
	
	\paragraph{Decidability of Non-monotonic SWRL Extensions:} As the authors of \cite{Calero2010} described, decidability issues involving unbound variables can be resolved through careful management of their syntax or semantics during the inference process. For example, in the case of the 'not' Operator --- negation as failure (see Table \ref{tab:weaknegation}). Here, decidability is achieved by controlling the appearance of the unbound variable in the consequent of the SWRL rule. In essence, the unbound variable (?z), whose inference can result in anonymous negative property assertions from the new elements not(P(x; y)) and not(Q(x; y)), was deliberately introduced so as not to use the bounded variables '?x' or '?y' in the consequent of the extended SWRL syntax. In effect, while the negation of the object or datatype property assertions P or Q holds, the resulting unbound variable (?z) that hold this momentary value will not appear in any other case or rule atoms and therefore the knowledge base remains decidable. \par 
	Similarly, in order to preserve the decidability of the main ontology while using the proposed NotExists ($\nexists$) quantifier (see Table \ref{tab:notExists}), the unbounded variable introduced is semantically defined to match all the individual IDs available in the ontology. In other words, the free variable is interpreted to be 'universally satisfiable'. This condition ensures that the execution of the operator does not result in any modification of the existing ontology and therefore remains decidable.\newline \newline
	Other forms of the non-monotonic SWRL extensions, such as the 'Mutex' and 'Dominance' operators, do not have a direct consequence on the content of the ontology and therefore may not affect its decidability or otherwise. However, the Mutex and Dominance operators, which controls and prioritizes the sequence of rule executions resp., can be used to control the Decidability of SWRL ontologies during the inference process. This is because, by enabling the Reasoner to block and/or schedule rule execution plan, they mimic the feature of the inference process. As such, undecidable rule fragments can be dominated or muted where necessary to achieve consistency.
	
	\paragraph{Decidability of Fuzzy SWRL Extensions:} Theoretically, the decidability of f-SWRL extension is not debatable considering that the extension simply introduces weighted values (see section \ref{fswrl}) to justify the importance of certain facts over others. Specifically, the class assertion, property assertion and rules axioms of SWRL were extended to show the degree of confidence of such assertions. As such, f-SWRL ontology --- and by extension the Vague-SWRL ontology, can be assumed decidable as long as their OWL class and property axioms conform to the DL-safety restrictions.
	While the semantics of the fuzzy axioms, such as class and property inclusion axioms may have different interpretations from the original SWRL, the concepts introduced in most of the fuzzy-based SWRL extensions reviewed, do not seem to invalidate the DL-Safeness principle and can therefore, be expected to remain decidable. 
	This can be seen from the fact that no new concepts or anonymous individuals are expected from these extensions nor the interpretation of their syntax --- which in most cases is closely similar to the original SWRL. For example, the comparable SWRL-F extension (see section \ref{swrlf}) introduces the concept of fuzzy ontology as a separate entity to maintain consistency. As such its inference has no direct influence to the consistency of the main ontology. Inherently, the Decidability of this extension also follows the DL-Safety restriction of the SWRL rules.
	
	\paragraph{Decidability of SWRL First Order Logic (FOL) Extensions:}
	FOL-based languages are generally undecidable and logical formalisms extending FOL, such as the SWRL-FOL, are usually so --- unless where they are restricted by some semantic or syntactic completeness theorems. 
	The SWRL FOL extension(see section \ref{swrlfol}) introduces assertion axioms that contain first-order formulae. However, the restriction is that quantified variables must be bound to their corresponding OWL typed quantifiers --- meaning that no 'free', 'anonymous' or 'unbounded' variables should exists in the rules. In such cases, the decidability of SWRL-FOL extension can also be said to be inherent in the DL-safety restriction of the component SWRL ontology. As the authors of SWRL introductory paper \cite{HORROCKS2005}, expressed regarding the semantic interpretation of FOL assertions; "An ontology is consistent if-and-only-if it is satisfied by at least one interpretation" Moreover, neither n-ary predicates nor functions were directly included in the SWRL-FOL abstract syntax as they do not fit with OWL and by extension the SWRL paradigm.
	
	\paragraph{Decidability of SWRL Built-in Extensions:} Built-in extensions basically means that the extension utilizes the abstract syntax and semantics of the SWRL rule language. As such, poses no decidability issues as long as the extensions are used within the DL-safety limits.\par 
	Most of the SWRL mathematical and temporal extensions were added as built-in predicates to the original SWRL definition in order to avoid the decidability and complexity overheads. Other cases, where major extensions were added, such as the 'OpenMath' extension discussed in section \ref{openmath}, the functionalities were strategically added based on a distinct separation between the mathematical and problem semantics. This is possible considering the fact that the mathematical functions operates on existing domain knowledge as input. Moreover, as the OpenMath extension also uses the built-in (\textit{swrlbext: mathext}) to define functions as instances of the $Formula$ class, which are then solved using a constraint solver. Due to the modularity, the main ontology is thus free of the inconsistencies that may arise during math operations. Hence the extension basically follows the decidability of basic SWRL built-ins, the DL-safety. 
	
	\section{Discussion and Summary}
	\label{discussionandsummary}
	As categorically summarized in Table \ref{tab:extensionssummary}, various extensions to SWRL syntax and semantics have been proposed and justified as necessary expressiveness extensions of the rule language. While the paper present them as six categories for clarity, the extensions can be basically summarized into four categories, viz. the (i) Uncertainty management extensions --- comprising of the fuzzy and probabilistic extensions, (ii) the Non-monotonic extensions --- comprising of the negation, quantifiers, rules ordering and prioritization, as well as the existential extensions, (iii) the Advanced mathematical extension featuring the SWRL-OpenMath extension, and lastly, (iv) the SWRL Built-in extensions.\newline
	\par
	In the fuzzy extension category, we discussed the SWRL-F, Vague-SWRL, F-SWRL, and FNSWRL, all of which were aimed at extending SWRL to enable the representation of uncertainties or incomplete information. The extensions basically involve the use of fuzzy logic principles, specifically the fuzzy and vague sets, to extend the description logic of the SWRL formalism. Their similarity is apparent in their use of the fuzzy weight to denote the degree of membership for individual instances of the fuzzy class or property within a SWRL axiom. An exception to this similarity is the 'F-SWRL extension' --- which introduces a \textit{fuzzymatch} operator for representing uncertainties based on the principles of fuzzy control system. A probabilistic extension of the SWRL formalism, 'Bayes-SWRL', was also discussed as another uncertainty-handling or predictive modeling extension of SWRL. Bayes-SWRL uses the Bayesian Networks of probability theory to manage predictive knowledge modeling using SWRL rules. See section \ref{bayesian} for details.

	Various non-monotonic expressiveness extensions of SWRL have been geared towards solving the classical negation and the negation as failure (NAF). The negation extensions were included here only for completeness as the issues have been thoroughly addressed both in the mainstream OWL and the SWRL's abstract syntaxes. It is safe to assume that these numerous proposals with their justifications lead to the advancement of the abstract definitions to handle the knowledge negation. Other expressiveness extensions discussed in the non-monotonic category include the existential extensions --- introducing existential operators with syntax such as: the exclamation mark '!' in the 'X-SWRL extension', the existential ($\exists$) and Universal ($\forall$) quantifiers in section \ref{nonmonotonic}, the '$\epsilon$' operator in the SWRL epsilon extension, and the 'makeOWLThing' operator proposed in the SWRLX built-in definition. All these extensions were recognized to allow a safe creation of individual instances to an OWL class or property, except the 'makeOWLThing' built-in --- where the resulting instance is beyond the inference of existing OWL reasoners. \newline
	\par 
	As domain knowledge is ever-evolving, there is sometimes the need to control the execution of rules or even retract some facts based on new found information. These challenges were addressed through rules ordering (using the 'dominance' operation), rule exclusion (using the 'mutex' operation) and facts removal, using the notExists operator ($\nexists$). These extensions (See sections \ref{nonmonotonic}-\ref{existentials}), which obviously entails the essence of non-monotonicity, were thence summarized in Table \ref{tab:extensionssummary} as non-monotonic extensions of SWRL.
	
	Advanced mathematical extensions of the SWRL formalism through built-ins were also discussed. The category introduces a special built-in extension based on the OpenMath language syntax and semantics. The 'SWRL-OpenMath' extension involves the use of \textit{OpenMath} functionalities to enable expressing mathematical formulas and scientific equations in SWRL. Another more classical approach in this categoryis the 'SWRLM extension', which introduces advanced math operations, beyond those in the core SWRL definition. SWRLM extension handle operations such as the square-root (swrlm:sqrt), natural log (swrlm:ln), etc. Another important built-in extension is the SWRL Temporal extension, which do not directly falls under any of the above four categories, was also discussed to highlight how time-related domain information and temporal facts can be expressed using SWRL rules. See sections \ref{mathextenstion}-\ref{swrlbuiltins} for details. 
	We present a compacted summary of these extensions and their component syntax and semantics in Table \ref{tab:extensionssummary} below.
	
		\begin{center}
		\captionof{table}{Table of Summary for \texttt{SWRL Extensions}}
		\tablefirsthead{\hline	\textbf{SWRL \newline Extension} & \textbf{Added Syntax} & \textbf{Added Semantics} & \textbf{Extension Type} \\\hline \hline}
		
		\begin{supertabular}{p{2.6cm}|p{3.8cm}|p{3.8cm}|p{2.6cm}}
			%\caption{Table of Summary for \texttt{SWRL Extensions}}
			\label{tab:extensionssummary}
			% header and footer information
			%\hline
			
			Fuzzy-SWRL \newline(f-SWRL) & Fuzzy class assertion: C(x) * w.\newline
			Fuzzy property assertion: P(x,y) * w.\newline
			Where: w $\in [0, 1]$. &	Introduces a truth value 'w', to specify degree of confidence for individual membership in a Class or Property. \newline &	Fuzzy \newline Extension\\ \hline
			
			Vague-SWRL &	Introduces a second-degree weight, $w_2$ to f-SWRL syntax.
			& Denote second degree of membership: \newline 
			w and $w_2$ specifies the upper and lower bounds of membership intervals in a Class or Property.\newline 	& Fuzzy \newline Extension\\ \hline
			
			SWRL \newline Fuzzy \small{(SWRL-F)} &	Introduce a fuzzy matching operator: \newline  \textit{fuzzymatch}(?x, 'fuzzy-value') &	Matches corresponding fuzzy variables with designated fuzzy values from fuzzy sets. \newline &	Fuzzy \newline Extension\\ \hline
			
			Fuzzy Non-monotonic SWRL (f-NSWRL) &	Fuzzy weight (w), 'Not' and '$\neg$' operators.	& Same as the F-SWRL and Negation extensions. &	Fuzzy-Non-monotonic Extension.\newline \\ \hline
			
			Bayes-SWRL	& Probability weight (the p-variable, $p$)
			Class assertion: \newline C(x) * $p_x$. \newline
			Property assertion: \newline P(x, y) * $p_{xy}$.\newline
			Where: p  [0, 1].	& The p-variable ($p$) matches the probability of Class or Property assertions with predefined values in a conditional probability table (CPT).\newline &	Probabilistic Extension\\ \hline

			Negation	&The 'not' and '$\neg$' operators &
			Negation of existing concepts and asserting negative facts \newline & 	Non-monotonic  Extension\\ \hline
			
			SWRL Quantifiers &	Exists ($\exists_X$) operator \newline  &	Asserts new instances of the quantified variable x.\newline 
			&	Non-monotonic  \\ 
			
			& notExists ($\nexists_X$) operator in antecedent&Check for missing facts.\newline &\\
			& notExists ($\nexists_X$) operator in consequent &  Removal of Knowledge \newline &\\
			& forAll ($\forall_X$)&	Group-wise assertions \newline &\\ \hline
			
			Dominance &	Dominance operator: $dominance (R_x, R_y)$,\newline  R = rule identifier&	Implies: Rule $R_x$ has more priority in the execution order than $R_y$ & Rules \newline Ordering  \small{$(Priority)$} \\ \hline
			
			Mutex &	Mutex operator: $Mutex (R_x, R_y)$,  R = rule identifier &	Implies: Rule $R_y$ will not be executed in the event that  $R_x$ is already executed \newline &	Rule \newline Exclusion\\	\hline
			
			Extended SWRL (XSWRL) &	Existential operator'!' e.g. existentially quantified variable (!x) \newline &	Creates new individuals that satisfy the variable, x.	&  Non-monotonic Existential Extension\\ \hline
			
			SWRL-FOL/ CIF-SWRL & The 'ForAll ($\forall$)' and 'Exists ($\exists$)' operators &	Imposes first order logic (FOL) quantification operations on the SWRL variables \newline & FOL \newline Extensions\\ \hline
			
			Epsilon Extension &	The terminology constructor $\epsilon_X \phi(X)$ &	Creates new individuals as 'values of' class X. \newline &	Quantification extensions\\ \hline
			
			SWRL OpenMath  &	The math built-in: $swrlbext:mathext$ &	Uses the functionalities of OpenMath to create additional built-ins \newline	& Advanced Math Ext.\\ \hline
			
			SWRL Temporal & Duration operations, Allen's Temporal intervals, add/subtract operations e.g. \small{temporal:before($T_1, T_2$)} \newline	& Implements temporal operations as predicates on valid-time data & Built-in Extensions\\ \hline
			
			SWRLM & Complex Mathematical operations (evaluate, square-root, natural log)  e.g. \small{$swrlm:eval$(?area, "width * height", ?width, ?height)} \newline & Implements mathematical functions as predicates & Built-in Extensions\\ \hline
			
			SWRLX & Existential Built-in e.g. \small{$swrlx$:make-OWLThings(?x, ?y)} & Creates new individuals of type \small{$owl:Thing$} & Built-in Extensions\\ \hline
		\end{supertabular}
		
	\end{center}
	
	\section{Conclusion}
	\label{conclusion}
	In this paper, we presented the SWRL formalism, discussed its expressive limitations and then reviewed the expressiveness extensions proposed to manage those limitations. The review discussed the various fuzzy and probabilistic extensions of SWRL, non-monotonic extensions, existentials, advanced mathematical extensions, as well as the SWRL built-in extensions. We undertaken this review of common SWRL expressiveness extensions with the aim of asserting SWRL's overall expressivity and further highlighted its level of acceptance within the semantic web community. From the foregoing, it is evident that much of the available SWRL extensions have been directed towards solving three major issues, namely the uncertainties in domain knowledge, the non-monotonicity, and the representation of mathematical knowledge.\newline \newline	
	Based on our observation, usability of SWRL rules and its expressiveness extensions may be limited due to the following reasons: (i) Lack of readily-available reasoners that can draw inference on the new extensions within decidable portion of the language, (ii) Modeling real-world problems using SWRL is still at the research stage, with little large-scale developmental efforts, (iii) the availability of alternative rule and query languages such as SPARQL and RuleML that are able to achieve (usually in a turnaround fashion), some of the current limitations of SWRL. Though the RuleML is originally not designed to be a rule language, however, its adoption may have caused much decline to the use of SWRL.
	 
	In future works, we would like to further evaluate the practicalities of the many theoretical extensions reviewed and further validate the inference/decidability of the implemented SWRL extensions. Another research gap identified is the 'need for standardization'--- as some of the extensions can only be useful in localized application domains while few others require more justified syntax and semantics. It is hoped that-with the continuous development efforts on SWRL expressive powers and usability, it can evolve into a comprehensive and complete ontology rule language for the semantic web. \newline
	 
	\par {\bf Acknowledgement.} This work was partially supported by a Grant received through the CFFRC scheme, Govt. of Malaysia as project number CFFRCPLUS-CropB1-001.
	We also thank Dr. Natasha Alechina for her valuable and constructive comments on earlier drafts of the manuscript.

	%% The Appendices part is started with the command \appendix;
	%% appendix sections are then done as normal sections
	%% \appendix
	
	%% \section{}
	%% \label{}
	
	%% If you have bibdatabase file and want bibtex to generate the
	%% bibitems, please use
	%%
	%%  \bibliographystyle{elsarticle-num} 
	%%  \bibliography{<your bibdatabase>}
	
	\bibliographystyle{splncs03} 
	\bibliography{ref1}

\begin{thebibliography}{10}
\providecommand{\url}[1]{\texttt{#1}}
\providecommand{\urlprefix}{URL }

\bibitem{agarwal2005modeling}
Agarwal, S., Hitzler, P.: Modeling fuzzy rules with description logics. CEUR
  Workshop Proceedings  188(5) (2005)

\bibitem{Grigoris2008}
Antoniou, G., van Harmelen, F. (eds.): Semantic Web Primer - Second Edition.
  MIT Press, Cambridge, Massachusetts London, England (2008)

\bibitem{Baader2003DLH1}
Baader, F., Calvanese, D., McGuinness, D.L., Nardi, D., Patel-Schneider, P.F.
  (eds.): The Description Logic Handbook: Theory, Implementation, and
  Applications. Cambridge University Press, New York, NY, USA (2003)

\bibitem{Berners-Lee2009}
Berners-Lee, T.: {Semantic Web and Linked Data} (2009),
  \url{http://www.w3.org/2009/Talks/0204-campus -party-tbl/}

\bibitem{Boris2012}
Boris, M., Bernardo, C.G., Horrocks, I., Zhe, W., Achille, F., Carsten, L.:
  {OWL 2 Web Ontology Language Profiles ( Second Edition )}. W3C Recommendation
  (December),  1--43 (2012)

\bibitem{Calero2010}
Calero, J.M.A., Ortega, A., Perez, G.G.M., Blaya, J.A.B., Skarmeta, A.F.G.: {A
  Non-monotonic Expressiveness Extension on the Semantic Web Rule Language}.
  Journal of Web Eng {\ldots}  11(2)(0),  93--118 (2012),
  \url{http://dl.acm.org/citation.cfm?id=2230897}

\bibitem{Connor2011}
Connor, M.J.O.: {A Method for Representing and Querying Temporal Information in
  OWL}. Communications in Computer and Information Science  127 CCIS,(January
  2011),  97--110 (2011)

\bibitem{DaCostaPROWL}
Da~Costa, P.C.G., Laskey, K.B., Laskey, K.J.: Pr-owl: A bayesian ontology
  language for the semantic web. In: Proceedings of the 2005 International
  Conference on Uncertainty Reasoning for the Semantic Web - Volume 173. pp.
  23--33. URSW'05, CEUR-WS.org, Aachen, Germany, Germany (2005),
  \url{http://dl.acm.org/citation.cfm?id= 2889849.2889852}

\bibitem{Darwiche2010}
Darwiche, D.: {Bayesian networks}. Communications of the ACM  53(12),  80--90
  (2010)

\bibitem{Ding2006}
Ding, Z., Peng, Y., Pan, R.: {BayesOWL: Uncertainty modeling in semantic web
  ontologies}. Studies in Fuzziness and Soft Computing  204,  3--29 (2006)

\bibitem{Eiter:2008:CAS:1379469.1379712}
Eiter, T., Ianni, G., Lukasiewicz, T., Schindlauer, R., Tompits, H.: Combining
  answer set programming with description logics for the semantic web. Artif.
  Intell.  172(12-13),  1495--1539 (Aug 2008),
  \url{http://dx.doi.org/10.1016/j.artint. 2008.04.002}

\bibitem{elenius2012swrl}
Elenius, D., Riehemann, S.: Swrl-iq user manual (2012),
  \url{http://protegewiki.stanford.edu/images /5/57/SWRL-IQ\_manual.pdf}

\bibitem{Elenius}
Elenius, D., Stehr, M.o.: {Rules and Computation on the Semantic Web},
  \url{https://pdfs.semanticscholar.org/717c/
  fc3f88953e05e\\0b8d03d381d5fa50429f919.pdf}

\bibitem{gau1993vague}
Gau, W.L., Buehrer, D.J.: Vague sets. IEEE transactions on systems, man, and
  cybernetics  23(2),  610--614 (1993)

\bibitem{horn1951onsentences}
Horn, A.: {On Sentences Which are True of Direct Unions of Algebras}. J. Symb.
  Log.  16(1),  14--21 (1951)

\bibitem{Horrocks:2005:ORP}
Horrocks, I., Patel-Schneider, P.F., Bechhofer, S., Tsarkov, D.: Owl rules: A
  proposal and prototype implementation. Web Semant.  3(1),  23--40 (Jul 2005),
  \url{http://dx.doi.org/10.1016/j.websem. 2005.05.003}

\bibitem{HORROCKS2005}
Horrocks, I., Patel-schneider, P.F., Bechhofer, S., Tsarkov, D.: {OWL rules: A
  proposal and prototype implementation}. Web Semantics: Science, Services and
  Agents on the World Wide Web  3(1),  23--40 (jul 2005),
  \url{http://www.sciencedirect.com/science/ article/pii/\\S1570826805000053}

\bibitem{horrocks2004swrl}
Horrocks, I., Patel-Schneider, P.F., Boley, H., Tabet, S., Grosof, B., Dean,
  M., et~al.: Swrl: A semantic web rule language combining owl and ruleml. W3C
  Member submission  21, ~79 (2004)

\bibitem{Kifer2005reqs}
Kifer, M., Brook, S.: {Requirements for an Expressive Rule Language on the
  Semantic Web}. W3C Workshop on Rules Languages (April) (2005),
  \url{http://www.w3.org/2004/12/rules-ws/paper/34/}

\bibitem{kim2009semantic}
Kim, H.J., Kim, W., Lee, M.: Semantic web constraint language and its
  application to an intelligent shopping agent. Decision Support Systems
  46(4),  882--894 (2009)

\bibitem{Kim2003}
Kim, W., Lee, M., Hong, J., Wang, T., Kim, H.: {Merging Mathematical Constraint
  Knowledge with the Semantic Web using a Semantic Web Constraint Language}.
  Iconceptpress.Com  2003(October 2009) (2003),
  \url{http://www.iconceptpress.com/download/ paper/100618194339.pdf}

\bibitem{Klir1994fuzzysets}
Klir, G.J., Yuan, B.: Fuzzy Sets and Fuzzy Logic: Theory and Applications.
  Prentice-Hall, Inc., Upper Saddle River, NJ, USA (1995)

\bibitem{Li2011}
Li, W., Tian, S.: {XSWRL, an extended semantic web rule language and prototype
  implementation}. Expert Systems with Applications  38(3),  2040--2045 (mar
  2011), \url{http://linkinghub.elsevier.com/retrieve /pii/\\S0957417410007694}

\bibitem{Liu2013}
Liu, Y., Chen, S., Li, S., Wang, Y.: {Bayes-SWRL: A probabilistic extension of
  SWRL}. Proceedings - 9th International Conference on Computational
  Intelligence and Security, CIS 2013  1,  702--706 (2013)

\bibitem{mcguinness2004owl}
McGuinness, D.L., Van~Harmelen, F., et~al.: Owl web ontology language overview.
  W3C recommendation  10(10),  2004 (2004)

\bibitem{McKenzie2004a}
McKenzie, C., Gray, P., Preece, A.: {Extending SWRL to express fully-quantified
  constraints}. Lecture Notes in Computer Science (including subseries Lecture
  Notes in Artificial Intelligence and Lecture Notes in Bioinformatics)  3323
  LNCS,  139--154 (2004), \url{http://link.springer.com/chapter/10.10
  07/978-3-540-30504-0{\_}11}

\bibitem{mei2005reasoning}
Mei, J., Bontas, E.P.: Reasoning paradigms for swrl-enabled ontologies  (2005)

\bibitem{Motik2009}
Motik, B., Grau, B.C., Horrocks, I., Wu, Z., Fokoue, A., Lutz, C.: {OWL 2 Web
  Ontology Language Profiles} (2009), \url{http://www.w3.org/TR/owl2-profiles/}

\bibitem{O'Connor:2009:SQWRL}
O'Connor, M., Das, A.: Sqwrl: A query language for owl. In: Proceedings of the
  6th International Conference on OWL: Experiences and Directions - Volume 529.
  pp. 208--215. OWLED'09, CEUR-WS.org, Aachen, Germany, Germany (2009),
  \url{http://dl.acm.org/citation.cfm?id= 2890046.2890072}

\bibitem{Pan2006fswrl}
Pan, J.Z., Stoilos, G., Stamou, G., Tzouvaras, V., Horrocks, I.: f-swrl: a
  fuzzy extension of swrl pp. 28--46 (2006)

\bibitem{Paschke2010}
Paschke, A., Boley, H.: {Rule markup languages and semantic web rule
  languages}. In: Rule Markup Languages and Semantic Web Rule Languages, pp.
  1--24. IGI Global (2010)

\bibitem{patel2005proposal}
Patel-Schneider, P.F.: A proposal for a swrl extension towards first-order
  logic. W3C Member Submission, April  (2005),
  \url{http://www.w3.org/Submission/SWRL-FOL/}

\bibitem{patel2004proposal}
Patel-Schneider, P.F., Hayes, P., Horrocks, I., Harmelen, F.: A proposal for a
  swrl extension to first-order logic. Proposal, DARPA DAML Program  (2004),
  \url{http://www.daml.org/2004/11/fol/proposal}

\bibitem{Predoiu2007}
Predoiu, L.: {Probabilistic Models for the Semantic Web --- A Survey}.
  International Journal of Approximate Reasoning pp. 288--307 (2007)

\bibitem{Rosati2005b}
Rosati, R.: Semantic and computational advantages of the safe integration of
  ontologies and rules. Lect. Notes Comput. Sci. (including Subser. Lect. Notes
  Artif. Intell. Lect. Notes Bioinformatics)  3703 LNCS,  50--64 (2005)

\bibitem{sanchez2007extending}
S{\'a}nchez-Maci{\'a}n, A., Pastor, E., de~L{\'o}pez~Vergara, J.E., L{\'o}pez,
  D.: Extending swrl to enhance mathematical support. In: International
  Conference on Web Reasoning and Rule Systems. pp. 358--360. Springer (2007)

\bibitem{shen2007using}
Shen, G., Huang, Z., Zhu, X., Wang, L., Xiang, G.: Using description logics
  reasoner for ontology matching. In: Intelligent Information Technology
  Application, Workshop on. pp. 30--33. IEEE (2007)

\bibitem{TimBerners-Lee2011}
{Tim Berners-Lee}, J.H., Lassila, O.: {The Semantic Web}. Scientific American
  Magazine  (may 2011)

\bibitem{Wangfnswrl}
Wang, X., Ma, Z.M., Xu, C., Cheng, J.: Nonmonotonic fuzzy rules in the semantic
  web. In: 2009 Sixth International Conference on Fuzzy Systems and Knowledge
  Discovery. vol.~2, pp. 275--279 (Aug 2009)

\bibitem{Wang2008}
Wang, X., Ma, Z., Yan, L., Meng, X.: {Vague-SWRL: a fuzzy extension of SWRL}.
  International Conference on Web Reasoning and Rule Systems pp. 232--233

\bibitem{rulemlswrlx}
Wiki, R.: Swrl existentials, \url{http://wiki.ruleml.org/index.php/SWRL}

\bibitem{Wlodarczyk2010}
Wlodarczyk, T.W., Rong, C., O'Connor, M., Musen, M.: Swrl-f: A fuzzy logic
  extension of the semantic web rule language. In: Proceedings of the
  International Conference on Web Intelligence, Mining and Semantics. pp.
  39:1--39:9. WIMS '11, ACM, New York, NY, USA (2011),
  \url{http://doi.acm.org/10.1145/1988688.1988735}

\end{thebibliography}

	%% else use the following coding to input the bibitems directly in the
	%% TeX file.
	
	%\begin{thebibliography}{00}
	
	%% \bibitem{label}
	%% Text of bibliographic item
	
	%\bibitem{I. Horrocks, P. F. Patel-schneider, H. Boley, S. Tabet, B. Grosof, and M. Dean, “SWRL : A Semantic Web Rule Language Combining OWL and RuleML,” W3C Member submission 21. 2004.}
	
	%\end{thebibliography}
\end{document}